\documentclass{article}

\usepackage[accepted]{icml2019}
\usepackage[utf8]{inputenc}
\usepackage[T1]{fontenc}
\usepackage{upquote}
\usepackage{xcolor}
\usepackage{hyperref}
\usepackage{xspace}
\usepackage{url}
\usepackage{booktabs}
\usepackage{amsfonts}
\usepackage{amsmath}
\usepackage{nicefrac}
\usepackage{microtype}
\usepackage[nameinlink]{cleveref}
\usepackage{graphicx}
\usepackage{caption}
\usepackage{keystroke} 
\usepackage{ifthen}
\usepackage{soul} 
\crefname{equation}{}{}

\usepackage{etoolbox}
\makeatletter
\patchcmd\@combinedblfloats{\box\@outputbox}{\unvbox\@outputbox}{}{%
   \errmessage{\noexpand\@combinedblfloats could not be patched}%
}%
\makeatother

\newcommand{\KL}{\operatorname{KL}}
\newcommand{\E}[2]{\mathbb{E}_{#1}\left[#2\right]}
\newcommand{\clip}{\operatorname{clip}}

\newcommand{\boxedtext}[1]{\setlength{\fboxsep}{8pt}\noindent\fbox{\parbox{\dimexpr\linewidth-2\fboxsep-2\fboxrule}{#1}}}
\usepackage{array}
\newcolumntype{C}[1]{>{\arraybackslash}p{#1}}

\hypersetup{colorlinks,urlcolor=blue,linkcolor=blue,citecolor=blue,anchorcolor=blue,breaklinks=true}
\icmltitlerunning{Fine-Tuning Language Models from Human Preferences}
\begin{document}
\twocolumn[
\icmltitle{Fine-Tuning Language Models from Human Preferences}
\centering
\begin{minipage}{5in}
\begin{icmlauthorlist}
\icmlauthor{Daniel M.\ Ziegler$^*$\enspace}{}
\icmlauthor{Nisan Stiennon$^*$\enspace}{}
\icmlauthor{Jeffrey Wu\enspace}{}
\icmlauthor{Tom B.\ Brown\enspace}{} 
\icmlauthor{Alec Radford\enspace}{}
\icmlauthor{Dario Amodei\enspace}{}
\icmlauthor{Paul Christiano\enspace}{}
\icmlauthor{Geoffrey Irving\enspace}{}
\end{icmlauthorlist}
\end{minipage}
\vskip 0.5ex
OpenAI\\
\texttt{\{dmz,nisan,jeffwu,tom,alec,damodei,paul,irving\}@openai.com}
\icmlcorrespondingauthor{Paul Christiano}{paul@openai.com}
\vskip 0.3in
]
\printAffiliationsAndNotice{\textsuperscript{*}Equal contribution}

\begin{abstract}
Reward learning enables the application of reinforcement learning (RL) to tasks where reward is defined by human judgment, building a model of reward by asking humans questions.
Most work on reward learning has used simulated environments, but complex information about values is often expressed in natural language, and we believe reward learning for language is a key to making RL practical and safe for real-world tasks.  In this paper, we build on advances in generative pretraining of language models to apply reward learning to four natural language tasks: continuing text with positive sentiment or physically descriptive language, and summarization tasks on the TL;DR and CNN/Daily Mail datasets.  For stylistic continuation we achieve good results with only 5,000 comparisons evaluated by humans.  For summarization, models trained with 60,000 comparisons copy whole sentences from the input but skip irrelevant preamble; this leads to reasonable ROUGE scores and very good performance according to our human labelers, but may be exploiting the fact that labelers rely on simple heuristics.
\end{abstract}

\section{Introduction}

We would like to apply reinforcement learning to complex tasks defined only by human judgment, where we can only tell whether a result is good or bad by asking humans.  To do this, we can first use human labels to train a model of reward, and then optimize that model.  While there is a long history of work learning such models from humans through interaction, this work has only recently been applied to modern deep learning, and even then has only been applied to relatively simple simulated environments \citep{christiano2017deep,ibarz2018,bahdanau2018learning}.  By contrast, real world settings in which humans need to specify complex goals to AI agents are likely to both involve and require natural language, which is a rich medium for expressing value-laden concepts.  Natural language is particularly important when an agent must communicate back to a human to help provide a more accurate supervisory signal \citep{irving2018debate,christiano2018amplification,leike2018scalable}.

Natural language processing has seen substantial recent advances.  One successful method has been to pretrain a large generative language model on a corpus of unsupervised data, then fine-tune the model for supervised NLP tasks \citep{dai2015semi, peters2018deep, radford2018improving, khandelwal2019sample}.  This method often substantially outperforms training on the supervised datasets from scratch, and a single pretrained language model often can be fine-tuned for state of the art performance on many different supervised datasets \citep{howard2018universal}.  In some cases, fine-tuning is not required: \citet{radford2019language} find that generatively trained models show reasonable performance on NLP tasks with no additional training (zero-shot).

There is a long literature applying reinforcement learning to natural language tasks.  Much of this work uses algorithmically defined reward functions such as BLEU for translation \citep{ranzato2015sequence,wu2016google}, ROUGE for summarization \citep{ranzato2015sequence,paulus2017deep,wu2018learning,gao2019reward}, music theory-based rewards \citep{jaques2017sequence}, or event detectors for story generation \citep{tambwekar2018controllable}.  \citet{nguyen2017reinforcement} used RL on BLEU but applied several error models to approximate human behavior.  \citet{wu2018learning} and \citet{cho2019towards} learned models of coherence from existing text and used them as RL rewards for summarization and long-form generation, respectively.  \citet{gao2019preference} built an interactive summarization tool by applying reward learning to one article at a time.  
Experiments using human evaluations as rewards include \citet{kreutzer2018reliability} which used off-policy reward learning for translation, and \citet{jaques2019way} which applied the modified Q-learning methods of \citet{jaques2017sequence} to implicit human preferences in dialog.  \citet{yi2019towards} learned rewards from humans to fine-tune dialog models, but smoothed the rewards to allow supervised learning.  We refer to \citet{luketina2019survey} for a survey of RL tasks involving language as a component, and for RL results using transfer learning from language.  RL is not the only way to incorporate ongoing human feedback: \citet{hancock2019learning} ask humans what a dialogue system should have said instead, then continue supervised training.

In this paper, we combine the pretraining advances in natural language processing with human preference learning.  We fine-tune pretrained language models with reinforcement learning rather than supervised learning, using a reward model trained from human preferences on text continuations.  Following \citet{jaques2017sequence,jaques2019way}, we use a KL constraint to prevent the fine-tuned model from drifting too far from the pretrained model.  We apply our method to two types of tasks: continuing text in a way that matches a target style, either positive sentiment or vividly descriptive, and summarizing text from the CNN/Daily Mail or TL;DR datasets \citep{hermann2015teaching,volske2017tldr}.  Our motivation is NLP tasks where supervised data sets are unavailable or insufficient, and where programmatic reward functions are poor proxies for our true goals.  

For stylistic continuation, 5,000 human comparisons (each choosing the best of 4 continuations) result in the fine-tuned model being preferred by humans 86\% of the time vs.\ zero-shot and 77\% vs.\ fine-tuning to a supervised sentiment network.  For summarization, we use 60,000 human samples to train models that can roughly be described as ``smart copiers'': they typically copy whole sentences from the input, but vary what they copy to skip irrelevant initial text.  This copying behavior emerged naturally from the data collection and training process; we did not use any explicit architectural mechanism for copying as in \citet{see2017get,gehrmann2018bottom}.  
One explanation is that copying is an easy way to be accurate, given that we did not instruct labelers to penalize copying but do instruct them to penalize inaccuracy.
It may also reflect the fact that some labelers check for copying as a fast heuristic to ensure a summary is accurate.  Indeed, human labelers significantly prefer our models to supervised fine-tuning baselines and even to human-written reference summaries, but not to a lead-3 baseline which copies the first three sentences.

For summarization, we continue to collect additional
data and retrain our reward model as the policy improves
(\emph{online} data collection).
We also test \emph{offline} data collection where we train the reward model using
data from the original language model only;
offline data collection significantly reduces the complexity of the training process.
For the TL;DR dataset,
human labelers preferred the policy trained with online data collection
71\% of the time,
and in qualitative evaluations the offline model often
provides inaccurate summaries.
In contrast, for stylistic continuation
we found that offline data collection worked similarly well.
This may be related to the style tasks requiring very little data;
\citet{radford2017learning} show that generatively trained models can learn to classify sentiment from very few labeled examples.

In concurrent work, \citet{bohm2019rl}
also use human evaluations to learn a reward function for summarization, and optimize that reward function with RL.
Their work provides
a more detailed investigation
of the learned policy and reward function
on the CNN/Daily Mail dataset,
while we are interested in exploring
learning from human feedback more generally and at larger computational scale.
So we consider several additional tasks,
explore the effects of on-policy reward
model training and more data,
and fine-tune large language models for both reward modeling and RL.

\section{Methods}


\begin{figure}
\centering {\tiny Reward model training}
\centering \includegraphics[width=\columnwidth]{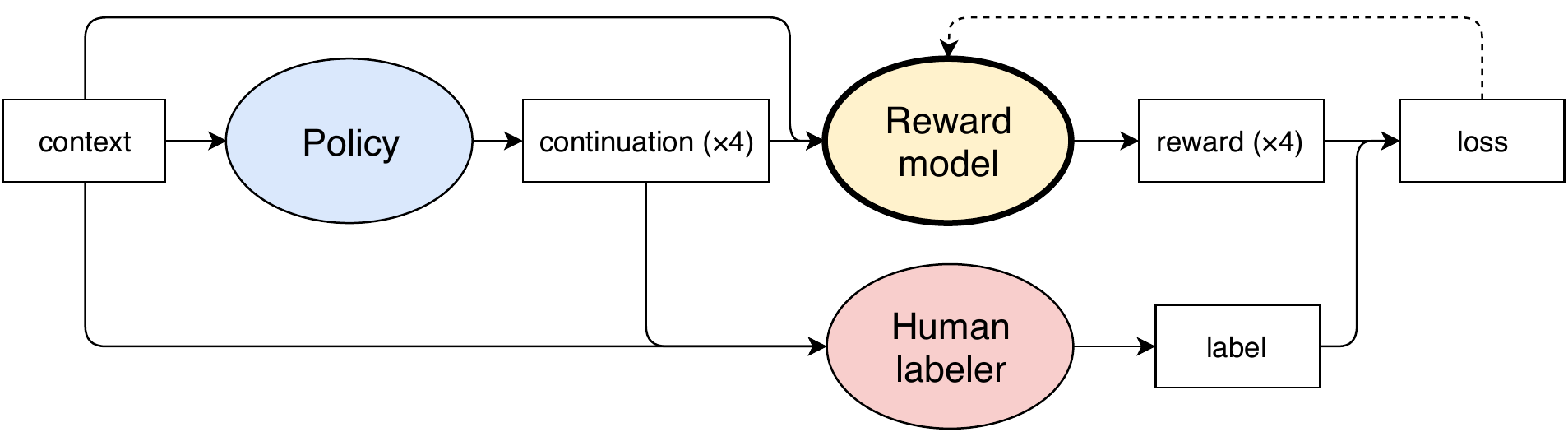}
\centering {\tiny Policy training}
\centering \includegraphics[width=\columnwidth]{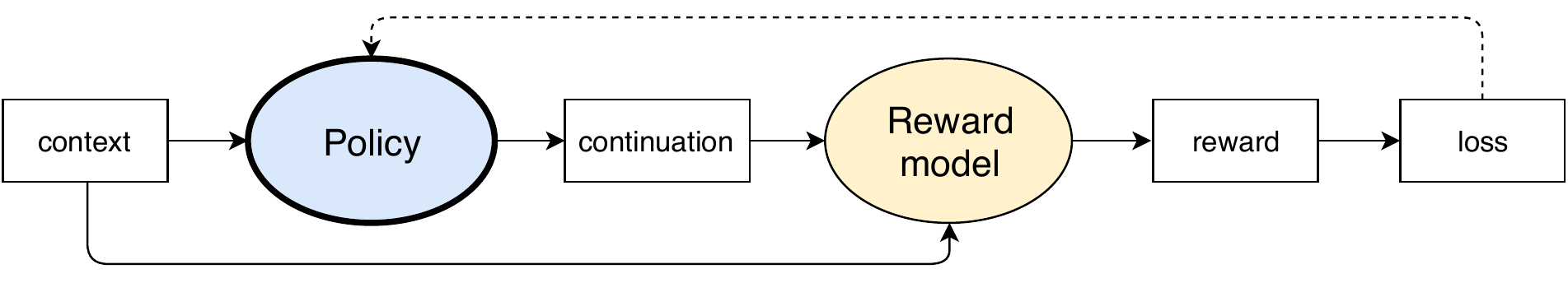}
\caption{Our training processes for reward model and policy.  In the online case, the processes are interleaved.} \label{fig:training-diagram}
\end{figure}

We begin with a vocabulary $\Sigma$ and a language model $\rho$ which defines a probability distribution over sequences of tokens $\Sigma^n$ via
\begin{align*}
\rho(x_0 \cdots x_{n-1}) = \prod_{0 \le k < n} \rho(x_k | x_0 \cdots x_{k-1})
\end{align*}
We will apply this model to a task with input space $X = \Sigma^{\leq m}$,
data distribution $\mathcal{D}$ over $X$,
and output space $Y = \Sigma^n$.
For example, $x \in X$ could be an article of up to 1000 words
and $y \in Y$ could be a 100-word summary.
$\rho$ defines a probabilistic policy for this task via
$\rho(y|x) = \rho(xy)/\rho(x)$:
fixing the beginning of the sample to $x$
and generating subsequent tokens using $\rho$.

\newcommand{\D}{\mathcal{D}}
We initialize a policy $\pi = \rho$,
and then fine-tune $\pi$ to perform the task well using RL.
If the task was defined by a reward function $r : X \times Y \rightarrow \mathbb{R}$,
then we could use RL to directly optimize the expected reward:
\[ \E{\pi}{r} = \E{x \sim \D, y \sim \pi(\cdot | x)}{r(x, y)} \]
However, we want to perform tasks defined by human judgments,
where we can only learn about the reward by asking humans.
To do this,
we will first use human labels to train a reward model,
and then optimize that reward model.

Following \citet{christiano2017deep}, we ask human labelers
to pick which of several values of $y_i$ is the best response to a given input $x$.%
\footnote{In early experiments we found that it was hard for humans to provide consistent fine-grained quantitative distinctions when asked for an absolute number,
and experiments on synthetic tasks confirmed that comparisons were almost as useful.}
We ask humans to choose between four options $(y_0, y_1, y_2, y_3)$; considering more options allows a human to amortize the cost
of reading and understanding the prompt $x$.
Let $b \in \left\{0, 1, 2, 3\right\}$
be the option they select.
Having collected a dataset $S$ of $(x, y_0, y_1, y_2, y_3, b)$ tuples,
we fit a reward model $r : X \times Y \rightarrow \mathbb{R}$
using the loss
\begin{equation}\label{eq:rploss}\textrm{loss}(r) = \E{\left(x, \left\{y_i\right\}_i, b\right) \sim S}{\log {\frac {e^{r(x, y_b)}}{\sum_i e^{r(x, y_i)}}}}\end{equation}
Since the reward model needs to understand language,
we initialize it as a random linear function of the final embedding output of the language model policy $\rho$
following \citet{radford2018improving} (see \cref{sec:sharing} for why we initialize from $\rho$ rather than $\pi$).
To keep the scale of the reward model consistent across training,
we normalize it so that it has mean 0 and variance 1 for $x \sim \D, y \sim \rho(\cdot|x)$.

Now we fine-tune $\pi$ to optimize the reward model $r$.
To keep $\pi$ from moving too far from $\rho$,
we add a penalty with expectation $\beta \KL(\pi,\rho)$ (see \cref{tab:samples-without-kl} for what happens without this).
That is, we perform RL on the modified reward
\begin{equation}\label{eq:rewarddef}
R(x, y) = r(x, y) - \beta \log \frac {\pi(y|x)}{\rho(y|x)}.
\end{equation}
We either choose a constant $\beta$ or vary it dynamically to achieve a particular value of $\KL(\pi,\rho)$; see \cref{sec:fine-tune}.
This term has several purposes: it plays the role of an entropy bonus,
it prevents the policy from moving too far from the range where $r$
is valid, and in the case of our style continuation tasks
it also is an important part of the task definition:
we ask humans to evaluate style,
but rely on the KL term to encourage coherence and topicality.

Our overall training process is:
\begin{enumerate}
\item Gather samples $(x,y_0,y_1,y_2,y_3)$ via $x \sim \D, y_i \sim \rho(\cdot | x)$.  Ask humans to pick the best $y_i$ from each.
\item Initialize $r$ to $\rho$, using random initialization for the final linear layer of $r$.  Train $r$ on the human samples using loss \cref{eq:rploss}.
\item Train $\pi$ via Proximal Policy Optimization (PPO, \citet{schulman2017proximal}) with reward $R$ from \cref{eq:rewarddef}
on $x \sim \D$.
\item In the online data collection case, continue to collect additional samples, and periodically retrain the reward model $r$.
This is described in \cref{sec:online}.
\end{enumerate}

\subsection{Pretraining details \label{sec:pretrain}}


We use a 774M parameter version of the GPT-2 language model in \citet{radford2019language} trained on their WebText dataset and their 50,257 token invertible byte pair encoding to preserve capitalization and punctuation \citep{sennrich2015neural}.  The model is a Transformer with 36 layers, 20 heads, and embedding size 1280 \citep{vaswani2017attention}. 

For stylistic continuation tasks we perform supervised fine-tuning of the language model to the BookCorpus dataset of \citet{zhu2015aligning} prior to RL fine-tuning; we train from scratch on WebText, supervised fine-tune on BookCorpus, then RL fine-tune to our final task.  To improve sample quality, we use a temperature of $T < 1$ for all experiments; we modify the initial language model by dividing logits by $T$, so that future sampling and RL with $T = 1$ corresponds to a lower temperature for the unmodified pretrained model.

\subsection{Fine-tuning details \label{sec:fine-tune}}

Starting with the pretrained language model, the reward model is trained using the Adam optimizer \citep{kingma2014adam} with loss \cref{eq:rploss}.  The batch size is 8 for style tasks and 32 for summarization, and the learning rate is $1.77\times 10^{-5}$ for both.  We use a single epoch to avoid overfitting to the small amount of human data, and turn off dropout.

For training the policy $\pi$, we use the PPO2 version of Proximal Policy Optimization from \citet{baselines}.  We use 2M episodes ($x,y$ pairs), $\gamma = 1$, four PPO epochs per batch with one minibatch each, and default values for the other parameters.  We use batch size 1024 for style tasks and 512 for summarization. We do not use dropout for policy training. The learning rate was $1.41\times 10^{-5}$ for style tasks and $7.07\times 10^{-6}$ for summarization.


Models trained with different seeds and the same KL penalty $\beta$ sometimes end up with quite different values of $\KL(\pi, \rho)$, making them hard to compare.  To fix this, for some experiments we dynamically vary $\beta$ to target a particular value of $\KL(\pi, \rho)$ using the log-space proportional controller
\begin{align*}
    e_t &= \clip\left(\frac{\KL(\pi_t, \rho) - \KL_{\mathrm{target}}}{\KL_{\mathrm{target}}}, -0.2, 0.2\right) \\
    \beta_{t+1} &= \beta_t( 1 + K_\beta e_t )
\end{align*}
We used $K_\beta = 0.1$.

For supervised fine-tuning baselines, we fine-tune for 1 epoch on the CNN/Daily Mail and TL;DR training sets (for TL;DR we removed 30K examples to serve as a validation set). We decayed the learning rate to 0 with a cosine schedule; for the initial value, we swept over 8 log-linearly spaced options between $10^{-4}$ and $3 \times 10^{-4}$. We also experimented with different dropout rates, and found a rate of 0.1 to work best.  We then chose the model with the best validation loss.  

\subsection{Online data collection \label{sec:online}}

If the trained policy $\pi$ is very different from the zero-shot policy $\rho$, the reward model will suffer a large distributional shift from training
on samples from $\rho$ to evaluation on samples from $\pi$.
To prevent this, we can collect human data throughout RL fine-tuning, continuously gathering new data by sampling from $\pi$ and retraining the reward model. 
As \cref{sec:experiments} shows, online data collection was important for summarization but not for the simpler style tasks.

In the online case,
we will choose a function $l(n)$ describing how many labels we want
before beginning the $n^{\textrm{th}}$ PPO episode.
Let $N_{\pi} = 2 \times 10^6$ be the total number of PPO episodes,
$N_r^0 = l(0)$ be an initial number of human labels,
and $N_r$ be the total number of human labels.
We take
\begin{align*}
l(n) &= N_r^0 + (N_r - N_r^0) \left(1 - (1 - n/N_\pi)^2\right)
\end{align*}
We pause before the $n^{\textrm{th}}$ PPO episode
if we have fewer than $l(n)$ labels.
We send another batch of requests to the labelers
if the total requests so far is less than $l(n) + 1000$,
to ensure they have at least 1000 outstanding queries at any time.
We train the reward model before the first PPO episode,
and then retrain it 19 more times at evenly spaced
values of $l(n)$.
Each time we retrain we reinitialize $r$ to a random linear layer on top of $\rho$
and do a single
epoch through the labels collected so far.
The offline case is the limit $N_r = N_r^0$.

To estimate overall progress, we gather validation samples consisting of $x \sim \D; y_0,y_1 \sim \rho(\cdot |x); y_2,y_3 \sim \pi(\cdot | x)$ at a constant rate; human labels on these give how often $\pi$ beats $\rho$.  Since validation samples are only used to evaluate the current $\pi$, we can add them to the training set for $r$. 
In order to estimate inter-labeler agreement, 5\% of queries are answered 5 times by different labelers.  Label counts in \cref{sec:experiments} include validation samples and repeated labels.

\subsection{Human labeling}

We use \href{https://scale.com}{Scale AI} to collect labels.
The Scale API accepts requests of the form $(x, y_0, y_1, y_2, y_3)$ and returns selections $b \in \left\{0, 1, 2, 3\right\}$.
We describe the task to Scale through a combination of instructions (\cref{sec:instructions})
and a dataset of about 100 example comparisons from the authors.

Unlike many tasks in ML,
our queries do not have unambiguous ground truth,
especially for pairs of similar outputs
(which play a large role in our training process,
since we train $r$ on pairs of labels sampled from a single policy $\pi$).
This means that there is significant disagreement even between labelers who
have a similar understanding of the task and are trying to rate consistently.
On 4-way comparisons for sentiment and TL;DR summarization, authors of this paper agree about 60\% of the time (vs. 25\% for random guessing).
This low rate of agreement complicates the quality control process for Scale;
the authors agree with Scale labelers 38\% of the time on sentiment and 46\% of the time on TL;DR summarization.
We give further details of the human data collection and quality evaluation in \cref{sec:qaprocess}.

For final evaluation of two models $A$ and $B$, we generate either 2-way comparisons between pairs $(a \sim A, b \sim B)$ or 4-way comparisons with quadruples $(a_0, a_1 \sim A, b_0, b_1 \sim B)$, randomize the order in which samples are presented, and present these comparisons to Scale.  Evaluating the quality of a model trained by Scale using the same set of humans from Scale is perilous: it demonstrates that $r$ and $\pi$ have succeeded in fitting to the human reward, but does not show that those human evaluations capture what we really care about,
and our models are incentivized to exploit idiosyncracies of the labeling process. 
We include samples from our models so that readers can judge for themselves.

\section{Experiments \label{sec:experiments}}

\begin{figure}
\centering \includegraphics[width=\columnwidth]{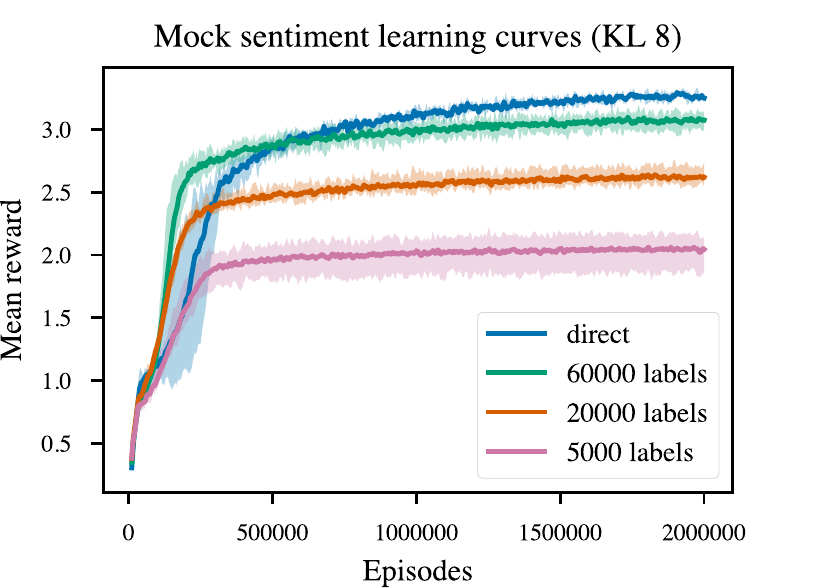}
\caption{Learning curves for a 124M-parameter model with mock sentiment reward, targeting a KL of 8 nats. Lines and shaded areas show mean and range for 5 seeds.  Early on the reward model sometimes speeds up training, a phenomenon also observed by \citet{christiano2017deep}.} \label{fig:mock-learning}
\end{figure}

\begin{figure*}
\vspace{-0.1in}
\centering \includegraphics[width=0.95\linewidth]{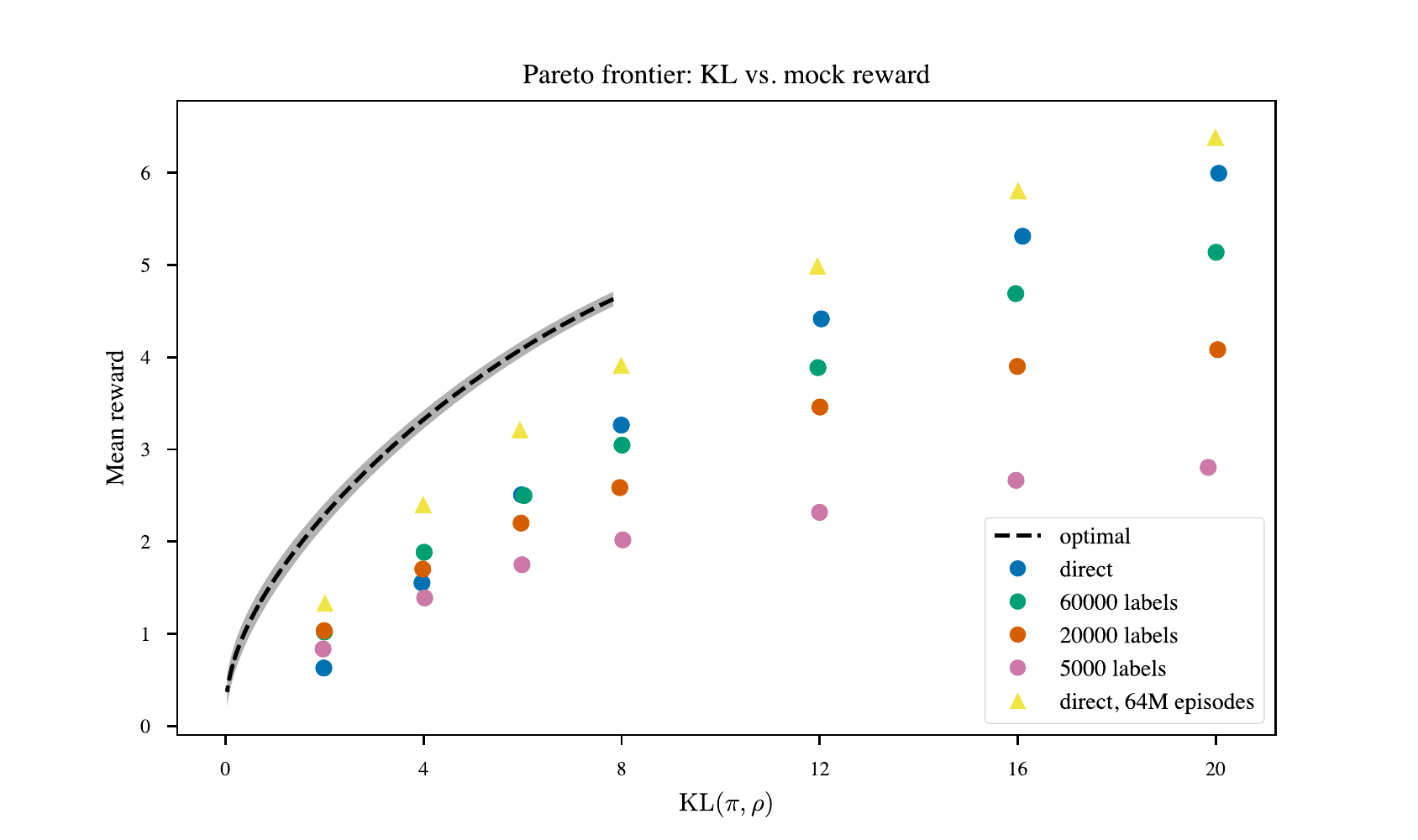}
\caption{Allowing the policy $\pi$ to move further from the initial policy $\rho$ as measured by $\KL(\pi,\rho)$ achieves higher reward at the cost of less natural samples.  Here we show the optimal KL vs.\ reward for 124M-parameter mock sentiment (as estimated by sampling), together with results using PPO.  Runs used 2M episodes, except for the top series.} \label{fig:kl-frontier}
\end{figure*}

In \cref{sec:mock}, we test our approach to RL fine-tuning of language models by using a mock labeler (a sentiment model trained on a review classification problem) as a stand-in for human labels.  We show that RL fine-tuning is effective at optimizing this complex but somewhat artificial reward.  In \cref{sec:style}, we show that we can optimize language models from human preferences on stylistic continuation tasks (sentiment and physical descriptiveness) with very little data, and that in the sentiment case the results are preferred to optimizing the review sentiment model. In \cref{sec:summarization} we apply RL fine-tuning to summarization on the CNN/Daily Mail and TL;DR datasets, show that the resulting models are essentially ``smart copiers'', and discuss these results in the context of other summarization work.

We \href{https://github.com/openai/lm-human-preferences}{release code}\footnote{Code at \href{https://github.com/openai/lm-human-preferences}{https://github.com/openai/lm-human-preferences}.} for reward modeling and fine-tuning in the offline data case.  Our public version of the code only works with a smaller 124M parameter model with 12 layers, 12 heads, and embedding size 768.  We include fine-tuned versions of this smaller model, as well as some of the human labels we collected for our main experiments (note that these labels were collected from runs using the larger model).

\subsection{Stylistic continuation tasks \label{sec:style-overview}}

We first apply our method to stylistic text continuation tasks,
where the policy is presented with an excerpt from the BookCorpus dataset \citep{zhu2015aligning}
and generates a continuation of the text.
The reward function evaluates the style of the concatenated text,
either automatically or based on human judgments.
We sample excerpts with lengths of 32 to 64 tokens,
and the policy generates 24 additional tokens.
We set the temperature of the pretrained model to $T = 0.7$ as described in \cref{sec:pretrain}.

\subsubsection{Mock sentiment task \label{sec:mock}}

\begin{figure*}[t]
\begin{minipage}[b]{.5\linewidth}
\centering \includegraphics[width=\columnwidth]{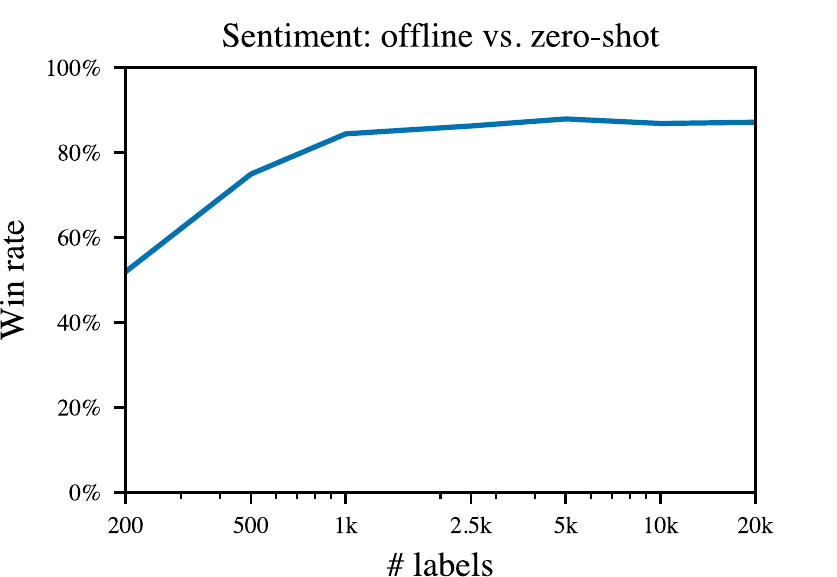}
\end{minipage}%
\begin{minipage}[b]{.5\linewidth}
\centering \includegraphics[width=\columnwidth]{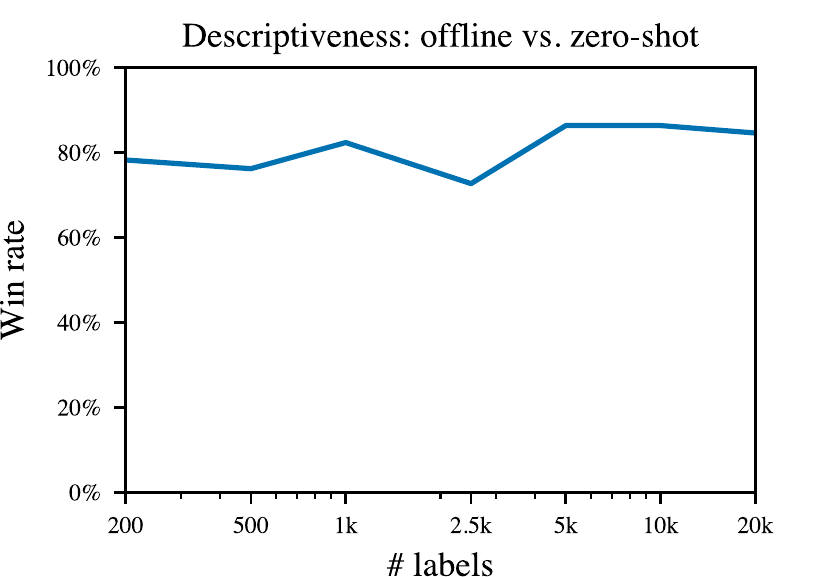}%
\label{fig:1b}
\end{minipage}
\caption{Human evaluations comparing the zero-shot model with offline fine-tuned models using varying amounts of human data.  We report how often the fine-tuned model is preferred by a majority of 3 labelers.  We omit error bars because we lack an estimate of the largest source of variance (randomness across training runs).
} \label{fig:style-eval}

\vspace{1em}
{

\newlength\MAX  \setlength\MAX{20mm}
\newlength\MAXPERC  \setlength\MAXPERC{0.2mm}
\newcommand*\Chart[4]{
    {\color{#3} \ifthenelse{#1 > \numexpr#2+6}{\textbf{#1\%}}{#1\%}}
    &
    \rlap{\textcolor{#4}{\rule{\MAX}{2ex}}}\textcolor{#3}{\rule{#1\MAXPERC}{2ex}}
    &
    {\color{#4} \ifthenelse{#2 > \numexpr#1+6}{\textbf{#2\%}}{#2\%}}
}

\newcommand{\tableentry}[5]{
  {\color{orange} 5k offline} vs. {\color{gray} #1}
  &  \Chart{#2}{#3}{orange}{gray}
  & \Chart{#4}{#5}{orange}{gray}
  \\}

\newcommand{\lefttableentry}[3]{
  {\color{orange} 5k offline} vs. {\color{gray} #1}
  & \Chart{#2}{#3}{orange}{gray}
  & \multicolumn{3}{c}{\large ---}
  \\}

\centering
\begin{tabular}{l|lll|lll} \toprule
  & \multicolumn{3}{c|}{Sentiment}
  & \multicolumn{3}{c}{Descriptiveness} \\ 
  \midrule

\tableentry{zero-shot}{88}{12}{86}{14}
\lefttableentry{mock}{77}{23}
\tableentry{20k offline}{48}{52}{47}{53}
\tableentry{5k online}{50}{50}{48}{52}

\bottomrule
\end{tabular}
\vspace{.1in}
\captionof{table}{Human evaluations for the sentiment and descriptiveness tasks.  We sample 1024 excerpts from the BookCorpus test set and report how often each model's continuations were preferred, as judged by a majority of 3 labelers.
\label{tab:human_evals_cont}}
} 
\end{figure*}

To study our method in a controlled setting,
we first apply it to optimize a known reward function $r_s$
designed to reflect some of the complexity of human judgments.
We construct $r_s$ by
training a classifier\footnote{The model is a Transformer with 6 layers, 8 attention heads, and embedding size 512.}
on a binarized, balanced subsample of the Amazon review dataset of \citet{mcauley2015image}.
The classifier predicts whether a review is positive or negative,
and we define $r_s(x, y)$ as the classifier's log odds that a review is positive
(the input to the final sigmoid layer).

Optimizing $r_s$ without constraints would lead the policy to produce incoherent continuations,
but as described in \cref{sec:fine-tune} we include a KL constraint
that forces it to stay close to a language model $\rho$
trained on BookCorpus.

The goal of our method is to optimize a reward function
using only a small number of queries to a human.
In this mock sentiment experiment,
we simulate human judgments by assuming that the ``human'' always selects the continuation
with the higher reward according to $r_s$,
and ask how many queries we need to optimize $r_s$.

\Cref{fig:mock-learning} shows how $r_s$
evolves during training, using either direct RL access to $r_s$ or a limited number of queries to train a reward model.
20k to 60k queries allow us to optimize
$r_s$ nearly as well as using RL to directly optimize $r_s$.

Because we know the reward function,
we can also analytically compute the optimal policy
and compare it to our learned policies.
\newcommand{\popt}[2]{\pi_{\textrm{opt}}(#1 #2)}
With a constraint on the KL divergence $\KL(\pi, \rho)$ between the learned policy $\pi$
and the language model $\rho$,
the optimal policy has the form:
\begin{align*}
\pi_\textrm{opt}(y | x) &\propto \rho(y | x) e^{r_s(x,y) / \beta }
\end{align*}
We approximate the reward of this policy for given $x$ and $\beta$
by sampling a large number of continuations from $\rho(y|x)$ and reweighting them by $e^{r_s(x,y) / \beta}$.
\Cref{fig:kl-frontier} compares the reward obtained by our policies
to the estimated optimal reward across a range of KL values.
There is a significant gap from optimality after training the policy
on 2M continuations---the number used in our main experiments---though it is largely closed with more training.
Our policies continue to receive higher rewards for larger KL divergences,
where we cannot afford to approximate $\pi_\textrm{opt}$ by sampling.

\subsubsection{Human evaluations of continuations\label{sec:style}}

\begin{table*}[t]
\begin{center}
\begin{tabular}{|p{2cm}|p{4cm}|p{4cm}|p{4cm}|}

\hline
\textbf{context}
&
\multicolumn{3}{|C{12cm}|}{ 
Pearl thought to herself that what they were about to do was exactly the sort of thing that they could do to help the villagers. They were all terrified of these guys.
\newline
At the police station the three walked up to the counter behind which was a senior constable studying some papers.
}
\\
\hline
&
\textit{Continuation 1}
&
\textit{Continuation 2}
&
\textit{Continuation 3}
\\
\hline
\textbf{zero-shot}
&
"Hello, I'm Pearl and this is my friend, Mike," said Pearl.
&
"May we speak to the police officer, sir?" asked the one in charge.
&
\textquotesingle{}Hello, can I help you?\textquotesingle{}
\newline
\textquotesingle{}Yes, we're the same people that the people were talking about.
\\
\hline
\textbf{5k offline fine-tune}
&
He turned to them with a smile. "Good afternoon, ladies. I'm Detective Inspector Jones.
&
The constable stood up and smiled as he saw them, obviously pleased to see them.
&
He smiled at them and waved them in, his eyes twinkling as he listened to their tales.
\\
\hline

\end{tabular}
\caption{Three random ($T=0.7$) continuations for our sentiment continuation task.  Chosen from appendix \cref{tab:samples-sentiment}; see appendix for more.}
\label{tab:samples-body-sentiment}
\end{center}
\end{table*}
    
\begin{table*}[t]
\begin{center}
\begin{tabular}{|p{2cm}|p{4cm}|p{4cm}|p{4cm}|}

\hline
    
\hline
\textbf{context}
&
\multicolumn{3}{|C{12cm}|}{ 
"I do not know if it was Viking related, but it could have been."
\newline
"Really?" Ailia said. Is it safe to be traveling here then?  Ailia looked behind her to make sure they weren't being followed.
}
\\
\hline
&
\textit{Continuation 1}
&
\textit{Continuation 2}
&
\textit{Continuation 3}
\\
\hline
\textbf{zero-shot}
&
There were no signs of anyone. 
\newline
"It is safe enough," Ailios said.
&
"Because I have a friend that is in the area and he will be coming with us.
&
It was hard to see that far.  "I do not like that word.
\\
\hline
\textbf{5k offline fine-tune}
&
Kaya crouched low, her eyes wide in the moonlight. Her body was tense.
&
She put her hand on the sword strapped to her back, and then pulled it out.
&
She strode out the door and walked down the street, her nose wrinkled in disapproval.
\\
\hline

\end{tabular}
\caption{Three random ($T=0.7$) continuations for our descriptiveness continuation task.  Chosen from appendix \cref{tab:samples-descriptiveness}; see appendix for more.}
\label{tab:samples-body-descriptiveness}
\end{center}
\vspace{-.1in}
\end{table*}
    


We apply our method to two continuation tasks defined by human judgments:
\begin{description}
\item[Sentiment:] Humans are asked to reward ``positive and happy'' continuations.
\item[Descriptiveness:] Humans are asked to reward ``vividly descriptive'' continuations.
\end{description}

The human labelers are presented with a BookCorpus excerpt and four possible continuations; they are asked to select the best continuation.  Full instructions for labelers are provided in \cref{sec:instructions}
(although labelers also learned from $\sim 50$ example comparisons labeled by the authors
and so the instructions do not completely define the task).

To make the labeling task more natural, we select excerpts that start and end with a period.
When sampling continuations that will be presented to humans,
we use rejection sampling to ensure there is a period between tokens 16 and 24
and then truncate at that period.\footnote{This is a crude approximation
for ``end of sentence.''
We chose it because it is easy to integrate into the RL loop,
and even a crude approximation is sufficient for the intended purpose
of making the human evaluation task somewhat easier.}
During the RL fine-tuning, we penalize continuations that don't have such a period
by giving them a fixed reward of $-1$.

We dynamically adjusted $\beta$ to obtain a KL divergence of 6 nats for descriptiveness
and 10 nats for sentiment (\cref{sec:fine-tune}).

\begin{table*}[t]
\centering
\begin{tabular}{l|cccc|cccc} \toprule
  & \multicolumn{4}{c|}{TL;DR} & \multicolumn{4}{c}{CNN/Daily Mail} \\ 
  & R-1 & R-2 & R-L & R-AVG & R-1 & R-2 & R-L & R-AVG \\ 
  \midrule

SOTA
& 22* & 5* & 17* & 14.7*
& 41.22 & 18.68 & 38.34 & 32.75 \\
\midrule
lead-3 
& 17.435 & 3.243 & 14.575 & 11.751
& \textbf{40.379} & \textbf{17.658} & 36.618 & 31.552
\\

zero-shot 
& 15.862 & 2.325 & 13.518 & 10.568
& 28.406 & 8.321 & 25.175 & 20.634
\\


supervised baseline 
& 17.535 & 3.124 & 14.969 & 11.877 

& 39.525 & 16.992 & 36.728 & 31.082
\\

supervised + 60k fine-tune
& \textbf{18.434} & \textbf{3.542} & \textbf{15.457} & \textbf{12.478}
& 40.093 & 17.611 & \textbf{37.104} & \textbf{31.603}
\\

60k fine-tune
& 16.800 & 2.884 & 14.011 & 11.232 
& 37.385 & 15.478 & 33.330 & 28.731
\\

30k fine-tune
& 16.410 & 2.920 & 13.653 & 10.994
& 35.581 & 13.662 & 31.734 & 26.992
\\

15k fine-tune
& 15.275 & 2.240 & 12.872 & 10.129
& 38.466 & 15.960 & 34.468 & 29.631
\\


60k offline fine-tune
& 16.632 & 2.699 & 13.984 & 11.105
& 33.860 & 12.850 & 30.018 & 25.576
\\





\bottomrule
\end{tabular}
\caption{ROUGE evaluations of summarization models.  For all models (excluding the lead-3 baselines), we sample with temperature 0.7 for TL;DR and 0.5 for CNN/Daily Mail.  We use the CNN/DM test set, but our own validation set for TL;DR. CNN/Daily Mail SOTA is from \citet{gehrmann2018bottom}.  * TL;DR SOTA is from \citet{gehrmann2019generating}, but the numbers are not comparable as we lack test set access and the TL;DR leaderboard uses an unofficial implementation of ROUGE.
\label{tab:rouge}}
\end{table*}
{
\setlength\MAX{20mm}
\setlength\MAXPERC{0.2mm}
\newcommand*\Chart[4]{
    {\color{#3} \ifthenelse{#1 > #2}{\textbf{#1\%}}{#1\%}}
    &
    \rlap{\textcolor{#4}{\rule{\MAX}{2ex}}}\textcolor{#3}{\rule{#1\MAXPERC}{2ex}}
    &
    {\color{#4} \ifthenelse{#2 > #1}{\textbf{#2\%}}{#2\%}}
}

\newcommand{\tableentry}[5]{
  {\color{orange} 60k fine-tuned} vs. {\color{gray} #1}
  &  \Chart{#2}{#3}{orange}{gray}
  & \Chart{#4}{#5}{orange}{gray}
  \\}

\newcommand{\alttableentry}[5]{
  {\color{blue} lead-3} vs. {\color{gray} #1}
  & \Chart{#2}{#3}{blue}{gray}
  & \Chart{#4}{#5}{blue}{gray}
  \\}

\begin{table*}[h!]
\centering
\begin{tabular}{l|rlr|rlr} \toprule
  & \multicolumn{3}{c|}{TL;DR}
  & \multicolumn{3}{c}{CNN/Daily Mail} \\ 
  \midrule


\tableentry{zero-shot}{96}{4}{91}{9}
\tableentry{supervised}{97}{3}{80}{20}
\tableentry{lead-3}{45}{55}{40}{60}
\tableentry{supervised + 60k fine-tuned}{80}{20}{74}{26}
\tableentry{30k fine-tuned}{40}{60}{62}{38}
\tableentry{15k fine-tuned}{79}{21}{47}{53}
\tableentry{60k offline fine-tuned}{64}{36}{65}{35}
\tableentry{reference summaries}{96}{4}{84}{16}

\midrule

\alttableentry{supervised}{97}{3}{89}{11}
\alttableentry{reference summaries}{97}{3}{89}{11}
\alttableentry{supervised + 60k fine-tuned}{75}{25}{85}{15}

\midrule

\bottomrule
\end{tabular}
\caption{Human evaluation of summarization models. For each pair of models and each dataset, we sample 1024 articles from the test set, generate a summary from each model, and ask 3 humans to pick the best summary using the same instructions as in training. The model chosen by a majority of the humans wins on that article. We report the fraction of articles that each model wins. For all models, we sample with temperature 0.7 for TL;DR and 0.5 for CNN/DM.
\label{tab:human_evals_summ}}
\end{table*}
} 




\begin{table*}[t]
\begin{center}
\begin{tabular}{|p{2cm}|p{12cm}|}

\hline
\textbf{reference summary}
&
Solar plane attempting to be first to circumnavigate world without using fuel is stuck in China. Solar Impulse 2 attempts to prove the power of renewable energy.\\
\hline
\hline
\textbf{zero-shot}
&
The plane has been grounded in China for two weeks because of bad weather.
\\
\hline
\textbf{60k fine-tune}
&
The Solar Impulse 2, the experimental plane attempting to fly around the world without using a drop of fuel, has been grounded by the weather in China.
\newline\newline
What was supposed to be an overnight pit stop in the southwestern city of Chongqing has now stretched into a two-and-a-half week stay.
\\
\hline
\textbf{supervised}
&
Solar Impulse 2 has been grounded by the weather in China. The plane took off from Abu Dhabi on March 9. The plane is trying to prove the power of renewable energy.
\\
\hline
\textbf{supervised + 60k fine-tune}
&
Solar Impulse 2 has been grounded in China for two-and-a-half weeks. The plane is attempting to fly around the world without using a drop of fuel. The team, which includes Bertrand Piccard, is taking turns flying the single-seater. The plane took off from Abu Dhabi on March 9 and has successfully flown through Oman.
\\
\hline

\end{tabular}
\caption{Random ($T=0.5$) summaries for our CNN/DM summarization task, on the same context.  Samples chosen from appendix \cref{tab:samples-summarization-cnndm} (see appendix also for context being summarized).  The 60k fine-tune model copies from the source article. }
\label{tab:summarization-body-samples}
\end{center}
\vspace{-.1in}
\end{table*}

We trained a range of models using different amounts of feedback,
testing both offline data collection where humans
rate only the initial language model's continuation,
and online data collection where humans continuously rate
the current policy's continuations (\cref{sec:online}).
We then compared these different policies to each other
and to the zero-shot performance of the original language model.
The results are shown in \cref{fig:style-eval} and \cref{tab:human_evals_cont}.
Each model comparison is based on 1024 four-way continuation comparisons,
two from each of the models being compared,
each rated by 3 humans.

For these continuation tasks, offline and online data collection give similar performance.
We find that very little human data is required for fine-tuning: performance with 5k, 10k, and 20k reward model training samples is similar, degrading only for less than 5k samples.%
\footnote{The descriptiveness policy trained with 2.5k samples performed poorly, but we believe this is due to randomness in RL.}
The model trained using the review sentiment classifier from \cref{sec:mock} does poorly relative to models optimized using human preference:
in 77\% of contexts, labelers preferred the output of the model trained with real human feedback.

\subsection{Summarization \label{sec:summarization}}

We also applied our method to two summarization tasks: the CNN/Daily Mail dataset of \citet{hermann2015teaching} and the TL;DR dataset of \citet{volske2017tldr}. We sample articles or Reddit posts, truncate to 500 tokens, add a \verb+"\n\nTL;DR:"+ suffix (and for CNN/Daily Mail, a \verb+"Article:\n\n"+ prefix) and let the policy respond with up to 75 tokens.  We set the temperature of the pretrained model to $T = 0.5$ for CNN/Daily Mail and $T = 0.7$ for TL;DR.  To make the task more natural for humans, we ensure articles consist of whole sentences by truncating to the last newline character.
When sampling summaries that will be shown to a human, we use rejection sampling to ensure
there is a newline between tokens 55 and 75 and truncate at that newline.
During RL fine-tuning, we penalize summaries that don't have such a newline
by giving them a fixed score of -1.
For CNN/Daily Mail we used a fixed KL coefficient $\beta = 0.1$; for TL;DR we used $\beta = 0.03$.

For RL fine-tuning, we trained online data collection models with 15k, 30k, and 60k human labels, and an offline data collection ablation with 60k labels.  We also show zero-shot performance of the pretrained model, a supervised fine-tuned baseline using the same pretrained model as starting point (\cref{sec:fine-tune}), and a lead-3 baseline which copies the first three sentences of the context.  We truncate lead-3 at a period in the same way we truncate generated summaries, so occasionally it is 2 sentences.  Finally, we combine supervised and RL fine-tuning: performing human RL fine-tuning starting with the supervised fine-tuned model.  The purely RL fine-tuned models use contexts from the datasets during training but ignore the reference summaries; the supervised and supervised+RL models use both contexts and summaries.

We report two sets of numerical results: human evaluations between pairs of models (\cref{tab:human_evals_summ}) and ROUGE results on the test set of CNN/Daily Mail and our validation set of TL;DR (\cref{tab:rouge}).  ROUGE results suggest that online data collection is important for best performance, in contrast to our stylistic continuation tasks.  At a fixed number of labels, online tends to be better than offline, with a 3 point R-AVG gain on CNN/DM at 60k labels.\footnote{That said, different training runs have considerable variation
and it is expensive to run multiple seeds with humans,
so it is possible that this gap is largely noise.}
On both datasets we see significant returns to data volume up to 60k human labels (though the trend is less clear for human evaluation).  On both datasets, supervised + RL fine-tuning is best, and indeed pure RL fine-tuning is worse than the supervised baseline according to ROUGE in all cases (though the supervised baseline uses the full supervised training dataset, which is much larger than 60k samples).  Lead-3 is hard to beat: it is the best model for R-1 and R-2 on CNN/Daily Mail, and only supervised + RL fine-tuning beats it otherwise.

But our goal is optimizing reward defined by humans, not ROUGE.
\Cref{tab:human_evals_summ} shows pairwise comparisons between different model pairs according to human labelers, using 1024 samples with majority vote of 3 labelers per sample. Here the picture is different, though also significantly noisier. Our online trained, 60k label model reliably beats both the zero-shot and supervised baselines, and even beats the combined supervised + RL fine-tuned model.  Online training remains important, but the situation w.r.t.\ data volume is less clear and likely contaminated by noise: the 60k TL;DR model beats the 30k model only 40\% of the time, for example.  More worrisome, the 60k online model beats the human ground truth 96\% of the time for TL;DR and 84\% of the time for CNN/Daily Mail.

What is going on?  As we show in the next section, our 60k RL fine-tuned model is almost entirely extractive (despite lacking any explicit extractive architectural component): it mostly copies whole sentences from the context, but varies which sentences are copied.

\subsubsection{What our models copy \label{sec:copiers}}

\begin{figure*}
\includegraphics[width=.8\columnwidth]{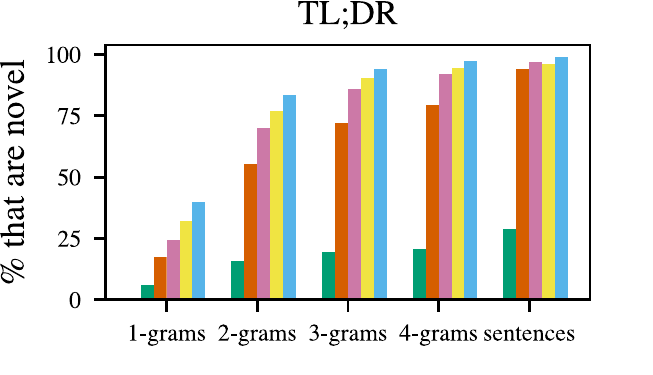}
\includegraphics[width=.8\columnwidth]{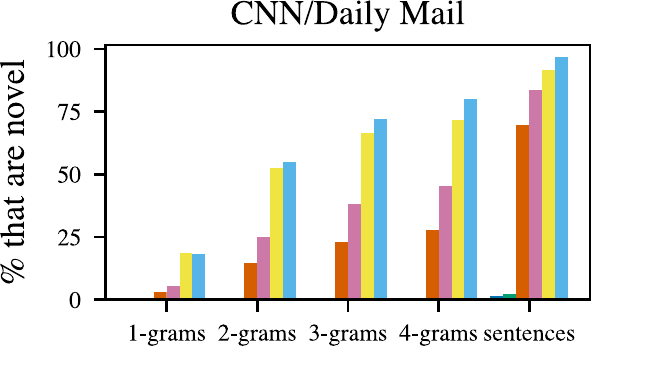}
\hspace{-0.05\columnwidth}
\raisebox{0.35\height}{\includegraphics[width=.45\columnwidth]{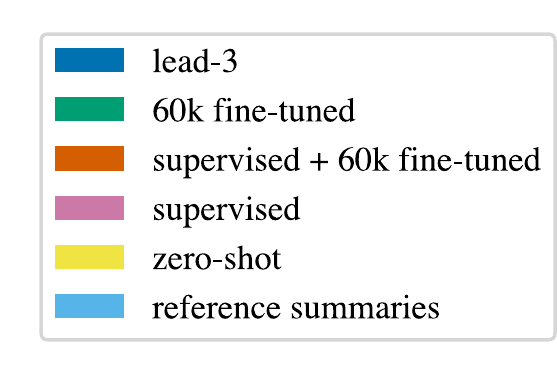}}
\caption{Percent of $n$-grams and sentences in summaries that do not appear in the source (compare to figure 6 in \citet{see2017get}). $n$-grams are consecutive sequences of words in a single sentence in a summary, and they count as novel if they do not appear consecutively in the article. We ignore punctuation and capitalization.}
\label{fig:copy-stats}
\end{figure*}

\begin{figure*}
\includegraphics[width=.8\columnwidth]{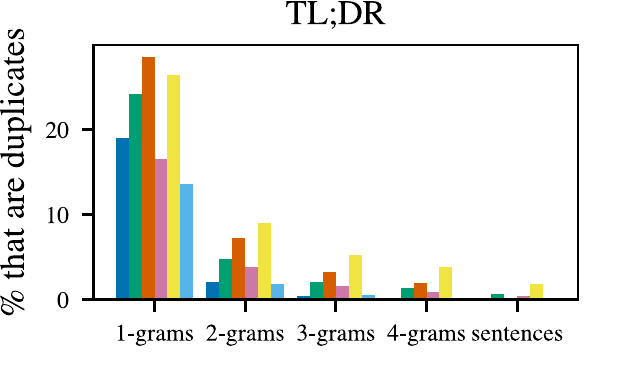}
\includegraphics[width=.8\columnwidth]{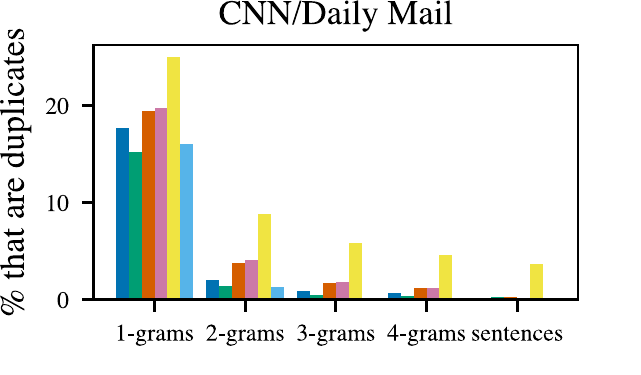}
\hspace{-0.05\columnwidth}
\raisebox{0.35\height}{\includegraphics[width=.45\columnwidth]{copy_stats_tldr_legend.pdf}}
\caption{Percent of $n$-grams and sentences in summaries that appear multiple times in the summary (compare to figure 4 in \citet{see2017get}).
}
\label{fig:duplicates}
\end{figure*}

Much previous work in summarization has focused on explicit copying mechanisms, including the pointer network-based architecture of \citet{see2017get} and the two-phase mask and paraphrase approach of \citet{gehrmann2018bottom}.  The goal is to take advantage of copying (which is of fundamental importance to the task of summarization) without only copying---to be abstractive rather than extractive.

\Cref{fig:copy-stats,fig:duplicates} show the fractions of $n$-grams and sentences generated by our models which are novel and repeated, respectively.  
From the novelty stats, we see that our RL fine-tuning consistently causes models to copy more.
In particular, our 60k RL fine-tuned models are almost entirely extractive: they copy whole sentences 71\% of the time for TL;DR and 98\% of the time for CNN/Daily Mail.  Applying RL fine-tuning starting from the supervised fine-tuned model copies much less: 6\% and 30\% for TL;DR and CNN/Daily Mail. 
Although we do not use explicit coverage metrics as in \citet{see2017get,gehrmann2018bottom}, both supervised and RL fine-tuned models do very little repetition within summaries.  

\begin{figure*}
\centering
\begin{minipage}[b]{.47\linewidth}
\centering {\small TL;DR}
\includegraphics[width=\columnwidth,height=1.5in]{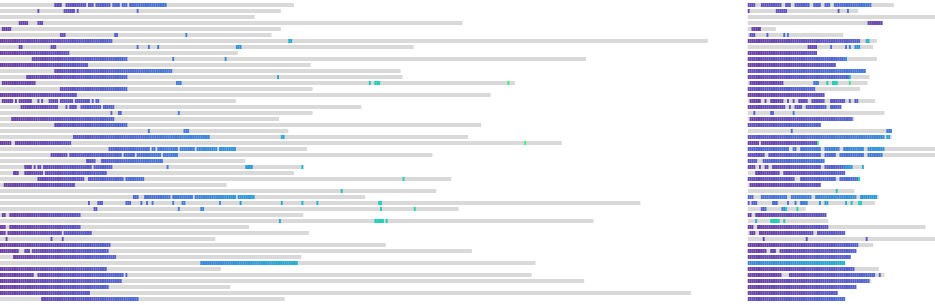}
\end{minipage}
\hspace{.3in}
\begin{minipage}[b]{.47\linewidth}
\centering {\small CNN/Daily Mail}
\includegraphics[width=\columnwidth,height=1.5in]{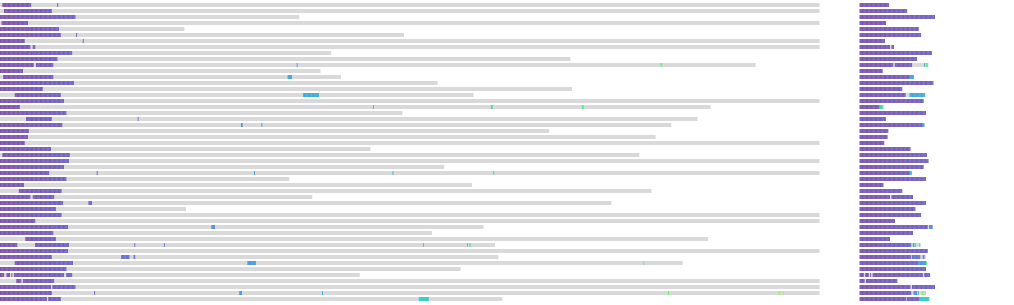}
\end{minipage}
\caption{Variation in where the models copy from, illustrated by the location of the longest common subsequence of bigrams between context article/post (left) and summary (right) for 256 randomly chosen contexts.  Document lengths are shown in gray, with bigrams highlighted (with color depending on positions in contexts).  Here, we picture the 60k fine-tuned models, which do the most copying.   \label{fig:heat-map}}

\end{figure*}

\begin{table*}[t]
\centering
\begin{tabular}{l|rr|rr} \toprule
  & \multicolumn{2}{c|}{TL;DR} & \multicolumn{2}{c}{CNN/Daily Mail} \\ 
  & all & preamble & all & preamble \\ 
  \midrule

zero-shot &
1.3\% & 0\% &
10.4\% & 1.0\% \\

60k fine-tuned &
28.3\% & 0.2\% &
77.6\% & 1.4\% \\

supervised &
1.5\% & 0\% &
9.4\% & 0\% \\

supervised + 60k fine-tuned &
7.9\% & 0\% &
16.6\% & 0\% \\

reference summaries &
0.6\% & 0\% &
5.1\% & 0\% \\

\midrule
total articles & 
30000 & 3762 &  
13368 & 297 \\

\bottomrule
\end{tabular}
\caption{How often different models copy the first 3 words of the article as the first 3 words of the summary, on the validation sets.  We additionally consider the subset of posts/articles with preambles.
On TL;DR, we used posts which begin with one of `hi', `hello', `hey', `ok', `okay', or `so'.  For CNN/Daily Mail, we used articles with a colon within the first 3 words, such as ``Winner: Simon Wood took home the TV crown [...]'' and ``Fully charged: The new scheme will let EE customers pick up free portable chargers [...]''.  \label{tab:copy-beginning-stats}}
\end{table*}

While the purely RL fine-tuned models mostly copy, they vary where they copy from. \Cref{fig:heat-map} illustrates this via the position of the longest common subsequence between context and summary.  To understand when the model chooses to copy from the exact beginning, we identify common preambles in articles such that we would expect copying to be a poor strategy.  \Cref{tab:copy-beginning-stats} shows that these preambles are copied much less often than in the immediate beginnings of other articles, giving evidence that our models are smart about when to copy.  However, we cannot establish that our reward model is smart beyond rewarding copying, as the zero-shot model also skips preambles.

\begin{table*}[t]
\centering
\begin{tabular}{l|r|r} \toprule
  & TL;DR & CNN/Daily Mail \\ 
  \midrule

zero-shot
& 6/30 & 6/30
\\

60k fine-tuned
& 26/30 & 29/30
\\

supervised
& 8/30 & 19/30
\\

supervised + 60k fine-tuned
& 11/30 & 20/30
\\

\bottomrule
\end{tabular}
\caption{Frequency with which generated summaries are accurate, in the sense of only making statements supported by the context, as judged by the authors on 30 articles from each dataset. 
The 60k fine-tuned model achieves high accuracy via copying;
the supervised and supervised + 60k fine-tuned models are more abstractive but at significant cost to accuracy.
\label{tab:accuracy}}
\end{table*}

Since combining supervised fine-tuning and RL fine-tuning gives the best ROUGE scores and and is also more abstractive, why not use it?  Unfortunately there is an advantage to pure copying shown in \cref{tab:accuracy}: it makes it easy for the model to tell the truth.  The models that copy the most, 60k RL fine-tuned, is 90\% and 95\% accurate on TL;DR and CNN/Daily Mail; lifting whole sentences from the article usually leaves them true.  The supervised fine-tuned and combined supervised+RL fine-tuned models are accurate at most 70\% of the time: they paraphrase but paraphrase badly, often swapping names from the context or mixing together multiple sentences in invalid ways.  Zero-shot is the most novel, but is accurate only 20\% of the time.  Similarly, \citet{kryscinski2019neural} found that 30\% of samples from the supervised summarization models they tested contained inconsistencies, and \citet{khandelwal2019sample} found that their pretrained encoder-decoder model ``hallucinates facts...which are topical but never appear in the source''.

There are at least two ways of interpreting these results.  The first is that copying is the easiest way to be accurate.  The labelers were told to penalize inaccuracy and redundancy, but were not told to penalize copying.  The zero-shot model copies some of the time, and when it copied it was accurate, so this behavior was reinforced.  The result is a model that ``degenerated to copying'', but at least does not lie.

However, this does not explain why both our model and lead-3 are strongly preferred by the labelers to the human reference summaries (\cref{tab:human_evals_summ}). This reveals a mismatch between the notion of quality we wanted our model to learn,
and what the humans labelers actually evaluated. Checking for copying is very easy, so labelers who check primarily for copying can work quickly.  Since the online data collection setting made quality control more difficult, we failed to detect and penalize this behavior.

\section{Challenges \label{sec:challenges}}

We conclude with a few lessons and directions we plan to consider in future reward learning work.

\subsection{Online data collection is hard}

Online data collection was necessary to achieve the best results on summarization.  However, fully online data collection---where each label comes from an up-to-date
version of the policy which has already learned
from almost all previous labels---had major disadvantages:
\begin{enumerate}
\item \textbf{Software complexity:} Our online system interleaves data gathering, reward model training, and RL fine-tuning.  The resulting distributed system was significantly more complicated than if each task was kept separate, slowing the software development process.  Moreover, a bug in any one of the tasks would break the entire training process. 
\item \textbf{Machine learning complexity:} Online experiments were difficult to debug, as it was hard to iterate on one piece of the ML system at a time.  We could often debug an online job only by switching briefly to offline, such as by launching a standalone reward model training run, but then would switch back to online once debugging was complete (until the next cycle).
\item \textbf{Quality control issues}: Significant work was required on Scale's part to make their data quality mechanisms work in the low latency, online setting.  However, even after this work it was difficult to maintain high data quality over a long period of time, and regressions were often not detected until after (or well after) training runs were complete.  Since evaluation of labeler performance was online, by the time a worker was detected as poor some of their data might already been reported back and used for reward model training.
\end{enumerate}
We believe the right middle ground between offline and online data collection is batched data collection, and plan to use this setting in future work.  Collect a batch of data from the pretrained policy $\rho$, train the reward model $r$ on this batch, then fine-tune the policy $\pi$ with $r$ frozen.  Once complete, collect another batch of data sampled from $\pi$, and iterate.  The latency for each batch can be far longer than the online case, simplifying quality control.  As in the fully online setting, we can always retrain the reward model from scratch on all data collected so far; human data is expensive so the total volume will be low.  Removing the interleaved training of $r$ and $\pi$ simplifies software architecture and diagnosis of ML issues, and allows iteration on just one component (say $r$ in isolation) if problems occur.  \citet{li2016dialogue} reached similar conclusions in a restricted dialogue setting after validating in simulation that online and batched trained performed similarly.

Batched data collection is also a well-studied setting for active learning techniques.  Although we use RL to fine-tune the policy $\pi$, the human data is used only for supervised training of the reward model $r$.  Thus, any method for batch mode active learning of supervised models applies, using $\pi$ as the unlabeled data distribution for $r$.  Examples of such techniques include selecting batches based on entropy considerations \citep{guo2008discriminative}, gradient-based metrics \citep{huang2016active,ash2019deep}, or by attempting to distinguish labeled and unlabeled examples \citep{gissin2019discriminative}.

\subsection{Sharing parameters between reward model and policy causes overfitting \label{sec:sharing}}

Although the reward model and policy are both initialized to $\rho$, we train them as separate networks rather than a single shared network with multiple heads.
We might expect joint training to be helpful, effectively using RL as an auxiliary task to improve the reward model's performance.
Joint training is particularly appealing
because it could help the reward model stay strong enough
that the policy cannot exploit it.
Sharing could also improve computational efficiency,
by allowing the models to share activations rather than requiring two separate forward passes.

Despite several attempts, we were not able to make this idea work.
The problem comes from the massive imbalance of data:
we have at most 60k samples for the reward model,
but 2M episodes for the policy.
This makes it challenging to maintain performance on both tasks
without performing many epochs for the reward model and overfitting.
We hope that future work will overcome this challenge.

\subsection{Ambiguous tasks make labeling hard}

Evaluation of a summary is both subjective and multidimensional.  A single human labeler may have a clear notion of whether a given sample is separately accurate, grammatical, nonredundant, or covers all important topics; but in our experiments
a labeler will often be asked to choose between samples each of which has some deficiencies.
In choosing which of four samples is the best, a labeler must trade off between different desiderata.
This makes consistent labeling difficult for honest labelers (including the authors!), and makes it difficult to quickly detect problematic labelers.
It also makes the research more difficult to present and interpret: during our experiments we routinely checked the performance of models by having authors label results, since we knew the authors would attempt to do the task honestly, but were epistemically uneasy about reporting these numbers in the paper (\cref{tab:accuracy} is the one exception).

One could hope to cope with such ``noise'' by simply getting more labels and averaging them,
but this does not resolve all the practical difficulties with ambiguity.
When possible, it seems better to design less ambiguous labeling tasks that get at the same information.
For example, rather than asking a person to rate or compare summaries, we could ask for a verbal description of the problems with a summary, or a suggested correction.  If problems don't exist we are done; otherwise describing a problem does not require consistently picking the same most important problem.  Even if two people disagree on the most important problem, they may be more likely to agree that the other picked \emph{some} problem, and more agreement eases data quality control and the overall experimental process.

\subsection{Bugs can optimize for bad behavior}

One of our code refactors introduced a bug which flipped the sign of the reward.  Flipping the reward would usually produce incoherent text,
but the same bug also flipped the sign of the KL penalty. 
The result was a model which optimized for negative sentiment while still regularizing towards natural language.
Since our instructions told humans to give very low ratings to continuations with sexually explicit text,
the model quickly learned to output only content of this form,
regardless of how innocuous the starting point was.  This bug was remarkable since the result was not gibberish but maximally bad output.
The authors were asleep during the training process, so the problem was noticed only once training had finished.
A mechanism such as Toyota's \href{https://en.wikipedia.org/wiki/Andon_(manufacturing)}{Andon cord} could have prevented this, by allowing any labeler to stop a problematic training process.

\section{Conclusion}

We have demonstrated RL fine-tuning of language models to four NLP tasks: stylistic continuation with high sentiment or physically descriptive language, and summarization on the CNN/Daily Mail and TL;DR datasets.  Rather than building task-specific techniques, we achieve our results by straightforwardly applying reward learning to language generation.  We extend previous reward learning work with pretrained models and KL regularization to prevent the policy from diverging too far from natural language.

Our results are mixed.  On the continuation tasks we achieve good results vs.\ the zero-shot baseline as evaluated by humans with very few samples: 2.5k for sentiment and 5k for descriptiveness.  However, for both summarization tasks our policies are only ``smart copiers'' (extractive rather than abstractive): they copy from the input text but skip over irrelevant preamble.  The advantage of copying is truthfulness: by comparison the zero-shot and supervised models produce natural, plausible-looking summaries that are often lies.  We believe the limiting factor in our experiments is data quality, in particular exacerbated by the online data collection setting, and plan to ameliorate this with batched data collection in future.

We believe the application of human reward learning to natural language tasks is important both from a capability and safety perspective.  On the capability side, purely supervised training is insufficient to correct mistakes that arise when sampling from trained policies, and RL training to programmatic reward functions such as BLEU or ROUGE is insufficient: \citet{paulus2017deep} conclude that "optimizing for single discrete evaluation metric[s] such as ROUGE with RL can be
detrimental to the model quality."  Interactive tasks such as dialogue are particularly relevant: it is difficult to define the goal of a dialogue without the human participant, and the length of dialogue makes it more likely that supervised learned models will go off distribution.  In the supervised case NLP models are trained using human data; if we want RL fine-tuning we need human data too.

On the AI safety side, interactive communication between humans and ML models is a requirement for scalable reward learning methods such as amplification, debate, and recursive reward modeling \citep{christiano2018amplification,irving2018debate,leike2018scalable}, and natural language is how humans communicate complex ideas.  Although language models are unlikely to be ready for these tasks in their full generality, \citet{perez2019convince} demonstrates that debate already improves generalization for question-answering when debaters quote from a source text.
Using direct human preferences for language tasks is a step in the direction of scalable reward learning for language, and we believe further steps are possible.

\section*{Acknowledgments}

We thank Akshat Bubna, Shariq Hashme, and many others at Scale for their work on data collection, Shan Carter for help with visualizations, Scott Gray for help with low precision training, Shahbaz Syed for information about the TL;DR dataset, and Christine Payne, Miles Brundage, Jared Kaplan, Jan Leike, Ethan Perez, and Jelena Luketina for helpful comments on the paper.

\bibliography{references}

\appendix
\section{Instructions for labelers\label{sec:instructions}}

\subsection{Sentiment}
\boxedtext{
You will see snippets of text followed by several possible ways the text
could be continued. Choose the most positive and happy continuation:

\begin{itemize}
\item
  Consider all of the moods of the characters, the mood of the narrator,
  and the mood implied by the events and descriptions.
\item
  If a continuation contains sexually explicit or otherwise offensive
  content, treat it as strongly negative. Do not choose it even if it
  expresses a positive attitude.
\end{itemize}
}

\subsection{Descriptiveness}
\boxedtext{
You will see snippets of text followed by several possible ways the text
could be continued. Choose the most vividly descriptive continuation:

\begin{itemize}
\item
  Evaluate both on the \textbf{quantity} and on the \textbf{vividness}
  of physical details described.
\item
  The best continuations are full of details that give a strong sense of
  what the scene looks, sounds, or smells like.
\item
  Count only \textbf{physical details}, not details about abstract
  facts.
\end{itemize}
}

\subsection{Summarization: TL;DR}
\boxedtext{
You will see some text followed by several summaries. Please read the text and select the best summary.

A summary is good if it:
\begin{itemize}
    \item Is useful and a good summary in general
    \item Accurately states the important points of the text
    \item Makes sense on its own
\end{itemize}
    
A summary is bad if it:
\begin{itemize}
    \item Includes information that doesn't appear in the text
\end{itemize}
}

\subsection{Summarization: CNN/DM}
\boxedtext{
You will see an article followed by several summaries. Please read the article and select the best summary.

A summary is good if it:
\begin{itemize}
    \item Is useful and a good summary in general
    \item Accurately states the important points of the article
    \item Makes sense on its own
\end{itemize}

A summary is bad if it:
\begin{itemize}
    \item Includes information that doesn't appear in the article
    \item Includes quotations that don't appear verbatim in the article
\end{itemize}
}

\section{Human labeling details \label{sec:qaprocess}}

Our quality assurance process was handled by Scale AI, though Scale made significant changes to their usual quality systems in order to deal with subjective tasks
and provide very fast turnaround.
Since we initially believed online data collection would be crucial, even the offline experiments were collected with this fast turnaround
requirement. In the future we plan to use a more relaxed latency requirement.

The first step of data collection involves teaching the task to a small number of trusted Scale labelers by giving them a description of the task (\cref{sec:instructions}).  Scale uses these labelers to collect a large number of \emph{benchmark} data points where several trusted labelers agree (out of a large set of unlabeled data points from $\rho$).  During full data collection, Scale serves these benchmark data points to freelance workers alongside real unlabeled data for training (the two types of data are indistinguishable when $\pi = \rho$, though they do become distinguishable during training), maintaining a confidence model for the performance of each labeler on the benchmark distribution.  The probability of getting a benchmark vs.\ a real sample varies dynamically on factors such as confidence in the labeler to correctly label a certain category.  Freelancers who fail to perform well on benchmark tasks are filtered out.  Additionally, Scale makes ad-hoc improvements to quality control over time, sometimes validating quality by comparing to a small number of gold-standard labels from the authors.

\begin{table}[t]
\begin{tabular}{l|r|r}
\hline
$P(\mathrm{agreement})$        & \multicolumn{1}{l|}{Sentiment}  & \multicolumn{1}{l}{TL;DR} \\
\hline
\textit{Between random responses} & \textit{25\%\phantom{$\pm$0}} & \textit{25\%\phantom{$\pm$0}} \\
Between labelers               & 38$\pm$2\% & 46$\pm$2\% \\     
Between an author \& a labeler & 44$\pm$5\% & 38$\pm$5\% \\
Between authors                & 62$\pm$5\% & 61$\pm$5\% \\
\hline
\end{tabular}
\caption{Agreement probabilities for two tasks, i.e.\ the probability of both individuals choosing the same sample as best, out of 4.}\label{tab:agreement}
\end{table}

We evaluated the data quality after the fact on two of the tasks.  During all data collection, 5\% of queries were answered by 5 distinct labelers.  We sampled 100 of these queries (restricting to ones generated from $\rho$) and had two authors label each one.  Based on this data, we estimated the rate of agreement between authors and Scale labelers, pairs of labelers, and pairs of authors.  As \cref{tab:agreement} shows, the data contained a significant amount of signal but did not match the quality of data which was hand-labeled by the authors.  

An earlier version asked labelers for 1-10 ratings; in the best case this provides more information per label, but it was difficult to gauge labeler performance.  Normalization was required since two good labelers would often differ by a (noisy) monotonic transform.  If many scores concentrated on a few values (say 7 and 8) simple strategies could fool the filtering process.  Absolute scores also tended to drift over the training process, as labelers would adjust to the new distribution of samples from the changing policy.

Finding high-quality workers involves human answering quality control questions
which are not used in our experiments, and throwing away data from low-quality workers.
So the total human cost of experiments is somewhat higher than the number of labels
we actually use (which is what we report).
For a short training run this can easily dominate the actual label requirements,
though it can be amortized across several tasks by identifying consistently good workers.
For our longer training runs the additional number of labels was modest.
(None of these details are exposed to customers.)

\section{Samples}

Samples from our models are shown in the following tables:
\begin{itemize}
\item Mock sentiment continuation without a KL penalty: \cref{tab:samples-without-kl} 
\item Sentiment continuation: \cref{tab:samples-sentiment}
\item Descriptiveness continuation: \cref{tab:samples-descriptiveness}
\item TL;DR summarization: \cref{tab:samples-summarization-tldr,tab:samples-summarization-tldr2,tab:samples-summarization-tldr3}
\item CNN/Daily Mail summarization: \cref{tab:samples-summarization-cnndm,tab:samples-summarization-cnndm2,tab:samples-summarization-cnndm3} 
\end{itemize}

\begin{table*}[t]
\begin{center}
\caption*{Samples without a KL penalty}
\begin{tabular}{|p{2cm}|p{6cm}|p{6cm}|}

\hline
    
\hline

\textbf{context}
&
\multicolumn{2}{|C{12cm}|}{ 
And she would always love him for that. And for many other reasons. He had given her so much and she
}
\\
\hline
&
\textit{Continuation 1}
&
\textit{Continuation 2}
\\
\hline
\textbf{zero-shot}
&
had no idea how to release that gift, but she knew
&
had been unable to give it back.
\newline\newline
"You
\\
\hline
\textbf{no penalty / bonus}
&
! These These These sound flowed instantly easily easily easily easily!
&
! These These These sound flowed instantly easily easily easily easily!
\\
\hline
\textbf{entropy bonus}
&
initially initiallyprisingly easilyprisingly Liam! Love this ride instantly tightly
&
' equally particularlyprisinglyprisingly Liam Liam! Love this ride instantly
\\
\hline
\hline

\textbf{context}
&
\multicolumn{2}{|C{12cm}|}{ 
Palmer rolled her eyes. "Don't give me that. You didn't come here, wearing that
}
\\
\hline
&
\textit{Continuation 1}
&
\textit{Continuation 2}
\\
\hline
\textbf{zero-shot}
&
corset and corset and a blasted corset. If
&
old-fashioned, sexy little dress. You came here to
\\
\hline
\textbf{no penalty / bonus}
&
! These These These sound flowed instantly easily easily easily easily!
&
! These These These sound flowed instantly easily easily easily easily!
\\
\hline
\textbf{entropy bonus}
&
initially relativelyprisingly instantlyprisingly Liam Liam! Love this ride instantly
&
overwhelmingly Loganprisingly Liam instantly surprisingly! Love this ride instantly tightly
\\
\hline
\hline

\textbf{context}
&
\multicolumn{2}{|C{12cm}|}{ 
Is it conceivable that - I don't know - he somehow wasn't on the aircraft that exploded? Like maybe he
}
\\
\hline
&
\textit{Continuation 1}
&
\textit{Continuation 2}
\\
\hline
\textbf{zero-shot}
&
simply didn't hear me coming or he was just coming out
&
was just on the ground."
\newline\newline
"Yes, at
\\
\hline
\textbf{no penalty / bonus}
&
! These These These sound flowed instantly easily easily easily easily!
&
! These These These sound flowed instantly easily easily easily easily!
\\
\hline
\textbf{entropy bonus}
&
prisinglyprisingly instantly Liam instantly overwhelmingly! Love this ride instantly tightly
&
instantly initially overwhelmingly Chloe! Love this ride instantly instantly instantly tightly
\\
\hline

\end{tabular}
\caption{Samples from a model fine-tuned to mock sentiment without a KL penalty.  Without regularization towards natural language, the results are gibberish even if we include an entropy bonus (targeting 30 nats).  Both policies reach a mean score of around $+8.0$ according to the mock sentiment model (i.e. $99.97\%$ positive).}
\label{tab:samples-without-kl}
\end{center}
\end{table*}
    

\begin{table*}[t]
\begin{center}
\caption*{Samples for positive-sentiment continuation task}
\begin{tabular}{|p{2cm}|p{4cm}|p{4cm}|p{4cm}|}

\hline
    
\hline
\textbf{context}
&
\multicolumn{3}{|C{12cm}|}{ 
"He also doesn't have a backbone," Quinn continued as they walked. "Last year, I had a great story about Paul Gibson, who is now the chairman of the board of supervisors.
}
\\
\hline
&
\textit{Continuation 1}
&
\textit{Continuation 2}
&
\textit{Continuation 3}
\\
\hline
\textbf{zero-shot}
&
He was running for the supreme court and his campaign was violent. There was an incident at a press conference.
&
For weeks, I went around the county, telling everyone how amazing he was and how much I admired him.
&
He was one of the first teachers to lead by example. It was incredibly inspiring to see.
\\
\hline
\textbf{mock}
&
I was really impressed with how he handled the board, and I thought it was a great story.
&
He was a real hero to the people in town. He was a real hero.
&
But he just sat around and watched. I love that story. You should read it again.
\\
\hline
\textbf{5k offline fine-tune}
&
Paul Gibson is a very intelligent man. He's a good heart man. He has a good sense of humor.
&
Paul is a very nice guy, a true gentleman, and we all know he's a really cute guy.
&
He was a good man, with a great smile, a great sense of humor, a great family.
\\
\hline
\hline
\textbf{context}
&
\multicolumn{3}{|C{12cm}|}{ 
Pearl thought to herself that what they were about to do was exactly the sort of thing that they could do to help the villagers. They were all terrified of these guys.
\newline
At the police station the three walked up to the counter behind which was a senior constable studying some papers.
}
\\
\hline
&
\textit{Continuation 1}
&
\textit{Continuation 2}
&
\textit{Continuation 3}
\\
\hline
\textbf{zero-shot}
&
"Hello, I'm Pearl and this is my friend, Mike," said Pearl.
&
"May we speak to the police officer, sir?" asked the one in charge.
&
\textquotesingle Hello, can I help you?\textquotesingle 
\newline
\textquotesingle Yes, we're the same people that the people were talking about.
\\
\hline
\textbf{mock}
&
"I'm Detective Sergeant Lewis, this is Detective Sergeant Jones and this is Detective Sergeant Porter.
&
He looked up at the three with a friendly smile.
\newline
"Good day, gentlemen.
&
"We have a friend who is in trouble. He was kidnapped by these guys," the boy said.
\\
\hline
\textbf{5k offline fine-tune}
&
He turned to them with a smile. "Good afternoon, ladies. I'm Detective Inspector Jones.
&
The constable stood up and smiled as he saw them, obviously pleased to see them.
&
He smiled at them and waved them in, his eyes twinkling as he listened to their tales.
\\
\hline
\hline
\textbf{context}
&
\multicolumn{3}{|C{12cm}|}{ 
The room was warm and welcoming, with two hearths that burned vigorously. The Vikings pushed her to the floor at Gunnar's feet. Ailia looked up at Gunnar and rose to her feet as calmly as possible.
}
\\
\hline
&
\textit{Continuation 1}
&
\textit{Continuation 2}
&
\textit{Continuation 3}
\\
\hline
\textbf{zero-shot}
&
"I will not fight."
\newline
"You will fight for us," Gunnar said.
&
"I'll give you one last chance to tell me where that is," Bjorn said.
&
"I'm sorry. I didn't mean to…"
\newline
"Give me your hand." Gunnar told her.
\\
\hline
\textbf{mock}
&
"I love this. I love this whole room. I'm going to enjoy it.
&
"I am very pleased to meet you, and let me assure you, you will not regret this.
&
"Bravely done, Gunnar. Thank you." Ailia smiled at him.
\\
\hline
\textbf{5k offline fine-tune}
&
Gunnar smiled at Ailia and waited for her to smile back. She was smiling, she knew.
&
"You've been a true friend to my son, Gunnar. Thank you." She smiled.
&
Gunnar was smiling at her and his pretty, blue eyes were filled with contentment.
\\
\hline

\end{tabular}
\caption{Continuations for our positive-sentiment continuation task, with contexts randomly sampled from BookCorpus test set.  Random samples from pretrained, mock sentiment, and 5k offline descriptiveness models, using temperature $T=0.7$.}
\label{tab:samples-sentiment}
\end{center}
\end{table*}
    

\begin{table*}[t]
\begin{center}
\caption*{Samples for descriptiveness continuation task}
\begin{tabular}{|p{2cm}|p{4cm}|p{4cm}|p{4cm}|}

\hline
    
\hline
\textbf{context}
&
\multicolumn{3}{|C{12cm}|}{ 
"I do not know if it was Viking related, but it could have been."
\newline
"Really?" Ailia said. Is it safe to be traveling here then?  Ailia looked behind her to make sure they weren't being followed.
}
\\
\hline
&
\textit{Continuation 1}
&
\textit{Continuation 2}
&
\textit{Continuation 3}
\\
\hline
\textbf{zero-shot}
&
There were no signs of anyone. 
\newline
"It is safe enough," Ailios said.
&
"Because I have a friend that is in the area and he will be coming with us.
&
It was hard to see that far.  "I do not like that word.
\\
\hline
\textbf{5k offline fine-tune}
&
Kaya crouched low, her eyes wide in the moonlight. Her body was tense.
&
She put her hand on the sword strapped to her back, and then pulled it out.
&
She strode out the door and walked down the street, her nose wrinkled in disapproval.
\\
\hline
\hline
\textbf{context}
&
\multicolumn{3}{|C{12cm}|}{ 
"They took me to a cellar. An' it were full of girls, it were, all sewin'. At first I thought it weren't too bad. I don't mind work, really I don't. Mama said I was a good 'elper.
}
\\
\hline
&
\textit{Continuation 1}
&
\textit{Continuation 2}
&
\textit{Continuation 3}
\\
\hline
\textbf{zero-shot}
&
But the girls keep on sneakin' in 'em. An' one day, one of em found me.
&
But in the end, it weren't anything like she'd promised me. Some of them were worse'n me.
&
I was just lucky to get this job. But as I stayed there, I started to see things.
\\
\hline
\textbf{5k offline fine-tune}
&
But..." The girl hiccupped, then pressed her lips together and shook her head.
&
But then..." His voice trailed off and he shook his head, his eyes going distant, somber.
&
She said I could keep my eyes open and my head clear and make a good target.
\\
\hline
\hline
\textbf{context}
&
\multicolumn{3}{|C{12cm}|}{ 
She forced herself up and shifted closer to him until she was kneeling behind him. Following her instincts, she laid a splayed hand on his bare back. He tensed beneath her touch. Quietly, Tara found her words. "I will never lie to you again, Max.
}
\\
\hline
&
\textit{Continuation 1}
&
\textit{Continuation 2}
&
\textit{Continuation 3}
\\
\hline
\textbf{zero-shot}
&
I love you."
\newline\newline
He turned to her, his face a mask of pain.
&
You have to know that."
\newline\newline
"I know you will never lie to me again.
&
Whatever happens, I'll tell you the truth."
\newline\newline
"I'm sorry.
\\
\hline
\textbf{5k offline fine-tune}
&
I promise."
\newline\newline
His hands curled around her forearms, his fingers digging deep into her flesh.
&
I swear it."
\newline\newline
He exhaled heavily and hugged her close, pressing his lips to her forehead.
&
I love you."
\newline\newline
Max leaned his head back and closed his eyes. His breathing changed.
\\
\hline


\end{tabular}
\caption{Continuations for our descriptiveness continuation task, with contexts randomly sampled from BookCorpus test set.  Random samples from pretrained and 5k offline descriptiveness models, using temperature $T=0.7$.}
\label{tab:samples-descriptiveness}
\end{center}
\end{table*}
    

\begin{table*}[t]
\begin{center}
\caption*{Samples for TL;DR summarization task (part 1)}
\begin{tabular}{|p{2cm}|p{12cm}|}

\hline
    
\hline
\textbf{context post}
&
\multicolumn{1}{|C{12cm}|}{ 
Okay, this was actually going to go into a comment on youtube, but Emma said she wanted more traffic on reddit.
\newline\newline
I am a musician myself, have been for many years now. While Emma's music is not the most complicated technical death metal stuff out there, I love it dearly. Her music actually moves me. No matter what she plays, be it a cover or original stuff, her music *touches* me on a deep personal level. And I have absolutely no clue why. I think it's a combination of her voice, her lyrics and the overall sound design of her songs and covers.
\newline\newline
But this isn't always a good thing, I feel. I can't listen to her music for hours on end. Only always a few songs, nothing more. Her music destroys me emotionally. Okay, that's a bit harsh, but she can get me, a (nearly) adult on the verge of tears. I've gone through some shit in my life, who hasn't, and I feel like her music can actually get some of it out. I am incredibly thankful that she makes her music the way she makes it. I can honestly say that Emma Blackery is my favourite artist. Keep in mind that I listen to a lot of music, from crossover thrash metal over 80's pop, berlin polygenre to classic/romantic music.
\newline\newline
I am also very jealous of her ability. I play quite a few instruments myself (guitar, piano, bass, drums, synth, saxophone, percussion) and aim to study audio engineering. I can definitely say that music is my literal life and mean it. I've released my first full length album at age 20. Yet I am very jealous that she is as good as she is and I really wanna know what her secret is. I love and hate her music, does that make sense?
\newline\newline
Her music definitely makes me think, and that's what art is all about, right? Anyway, I'm gonna continue working on my cover of The Promise and try to figure out what it is she's doing. Thank you for reading this far.
}
\\
\hline

\textbf{reference summary}
&
Emma music gud yes very love magic
\\
\hline
\hline

\textbf{zero-shot}
&
I'm a musician myself, have been for many years now. And Emma Blackery is my favourite artist.
\\
\hline
\textbf{60k fine-tune}
&
I am a musician myself, have been for many years now. While Emma's music is not the most technical death metal stuff out there, I love it dearly. Her music actually moves me. No matter what she plays, be it a cover or original stuff, her music *touches* me on a deep personal level.
\\
\hline
\textbf{supervised}
&
I'm jealous of Emma Blackery's music and I have no clue why.
\\
\hline
\textbf{supervised + 60k fine-tune}
&
Emma Blackery is my favourite artist, and I love her music to death. I wish I could listen to her music all the time. However, I feel like her music destroys me emotionally. I feel like I need to enjoy her music for the human part of it, and not the technical part. She's my favourite artist.
\\
\hline

\end{tabular}
\caption{Random ($T=0.7$) summaries for our TL;DR summarization task, with contexts randomly sampled from TL;DR validation set.}
\label{tab:samples-summarization-tldr}
\end{center}
\end{table*}

\begin{table*}[t]
\begin{center}
\caption*{Samples for TL;DR summarization task (part 2)}
\begin{tabular}{|p{2cm}|p{12cm}|}

\hline
    
\hline
\textbf{context post}
&
\multicolumn{1}{|C{12cm}|}{ 
It really does. I did the first 5 seasons in a couple months, and ultimately I enjoyed them. The first three seasons cover just about all the really good stuff. Russell Edgington was, to me, the best supporting character in the show. He was everything that I valued about the show, which is to say, campy, violent, overly-gory fun. I went back to the show a little over a year later, and while S6 was tolerable with a few fun bits, S7 totally eroded. Tara's plot goes full retard in the last season, people die for the sake of drama and convenience or leave with no real reverence for their history. Suddenly no one can make a good decision, and there was a needless romance bait-and-switch that ruined what I felt the show was moving towards.
\newline\newline
What's terrible is that once you've done six seasons it's hard to walk away, but each episode of seven stings more than the last. You don't get that weird half-hollow/half-warm feeling you should when you finish a series of books or a long TV show. Outside of Pam and Erik, I didn't give two fucks about any character by the end.
}
\\
\hline

\textbf{reference summary}
&
Seasons one through three were a blast. Four was pretty meh, five and six weren't too bad, seven sucked massively.
\\
\hline
\hline

\textbf{zero-shot}
&
I finished S6 and still feel like I'm in the same place as I was with S5.
\newline\newline
Favorite Episodes:
\newline\newline
S1: "Tough Love"
\newline\newline
S2: "Crossover"
\newline\newline
S3: "The Party"
\\
\hline
\textbf{60k fine-tune}
&
I did the first 5 seasons in a couple months, and ultimately I enjoyed them. The first three seasons cover just about all the really good stuff. Russell Edgington was, to me, the best supporting character in the show. He was everything that I valued about the show, which is to say, campy, violent, overly-gory fun
\\
\hline
\textbf{supervised}
&
10/10 would watch again
\\
\hline
\textbf{supervised + 60k fine-tune}
&
The first two seasons of the show were enjoyable. The show's progress in the last three seasons is frustrating because it's so similar to the first three seasons, and it's so one-sided that there's no reason to keep going. But the show's greatest strength has always been its great writing, and it's giving me that same feeling again.
\\
\hline

\end{tabular}
\caption{Random ($T=0.7$) summaries for our TL;DR summarization task, with contexts randomly sampled from TL;DR validation set.}
\label{tab:samples-summarization-tldr2}
\end{center}
\end{table*}

\begin{table*}[t]
\begin{center}
\caption*{Samples for TL;DR summarization task (part 3)}
\begin{tabular}{|p{2cm}|p{12cm}|}

\hline
    
\hline
\textbf{context post}
&
\multicolumn{1}{|C{12cm}|}{ 
> you should be happy, lots of votes were counted twice in precincts that went to Clinton in a landslide, if they recount them by hand she will lose more votes, and probably a lot...
\newline\newline
You are assuming you understand what the problem here is and what the end result on the election was.  The article does not state where the problem lies.
\newline\newline
E.g. The machine shows a tally of 120 votes having been processed on it.  The actual reported vote tally to the election board was 100 votes.  The 20 vote differential on the machine is due to 20 times a ballot needing to be re-scanned due to error.  No double-counting of votes occurs but there is also a delta between what the machine says was recorded vs what was reported.
\newline\newline
In this case the final reported vote tallies may have been factually correct but there is no way to validate them against the machine record.
\newline\newline
To understand the problem an analysis and deconstruction of the machines and the election night events must be undertaken.  Find out the cause of the problem, if the problem caused votes to be over/under counted, or if the problem was solely with machine data that was not used for tabulation of results.
}
\\
\hline

\textbf{reference summary}
&
neither you nor I have any idea what the actual problem is so there is no basis for saying the problem favors either candidate at this time, or if either candidate would even be affected.
\\
\hline
\hline

\textbf{zero-shot}
&
Election officials did not count ballots in a timely manner, by hand, to meet the threshold of getting votes counted on election night.
\newline\newline
Voters were not eligible to vote on election night and were not aware of the problems.
\newline\newline
The election officials did not realize that their system was broken until very late in the night.
\\
\hline
\textbf{60k fine-tune}
&
The machine shows a tally of 120 votes having been processed on it.  The actual reported vote tally to the election board was 100 votes.  The 20 vote differential on the machine is due to 20 times a ballot needing to be re-scanned due to error.  No double-counting of votes occurs but there is also a delta between what the machine says
\\
\hline
\textbf{supervised}
&
Stop assuming the election was completely free of double-counting before you even begin to discuss the actual results.  For any result to be close, it must be based on the total vote totals reported to the election board and not just the machine results.  The election board did not report the machine results to the election board.
\\
\hline
\textbf{supervised + 60k fine-tune}
&
The reported vote tallies may have been factually correct but there is no way to validate them against the machine record.  The machine may have actually been accurate and the machine may have been faulty in other ways.  The election may have been considered fair but the machine may not have been.  The results are still in dispute.
\\
\hline


\end{tabular}
\caption{Random ($T=0.7$) summaries for our TL;DR summarization task, with contexts randomly sampled from TL;DR validation set.}
\label{tab:samples-summarization-tldr3}
\end{center}
\end{table*}
    

\begin{table*}[t]
\begin{center}
\caption*{Samples for CNN/DM summarization task (part 1)}
\begin{tabular}{|p{2cm}|p{12cm}|}

\hline
    
\hline
\textbf{context article}
&
\multicolumn{1}{|C{12cm}|}{ 
Google launched a new U.S. wireless service today that switches between Wi-Fi and cellular networks
\newline\newline
Google is already the world's most popular phone software provider, and a pay-TV operator - and now it wants to be your mobile network carrier.
\newline\newline
The company has unveiled a U.S. wireless service that switches between Wi-Fi and cellular networks to curb data use and keep phone bills low.
\newline\newline
The service, called \textquotesingle{}Project Fi,\textquotesingle{} debuted today, about two months after Google revealed its plans to expand its ever-growing empire into providing wireless connections for smartphones.
\newline\newline
Google is selling the basic phone service for \$20 a month and will only charge customers for the amount of cellular data that they use each month, instead of a flat rate.  
\newline\newline
Each gigabyte of data will cost \$10 a month. That means a customer could sign up for a plan offering three gigabytes of data and get \$20 back if only one gigabyte was used in a month.\newline\newline[...\textit{Truncated}...]\newline\newline Rather than building its own network, Google is leasing space on cellular towers built by Sprint and T-Mobile, which are hoping the deals will boost their profits without costing them too many customers tempted to defect to Project Fi.
}
\\
\hline

\textbf{reference summary}
&
Project Fi will be hosted through Sprint Corp and T-Mobile's networks. It costs \$20 for basic service and unused data is paid back to customer. The invitation-only service will work only on Nexus 6 phones in the US. Numbers will live in the cloud so users can talk on any connected tablet.\\
\hline
\hline

\textbf{zero-shot}
&
Project Fi is a new smartphone service that switches between Wi-Fi and cellular networks to curb data use and keep phone bills low. The service is available for an introductory price of \$20 a month and will only charge customers for the amount of cellular data that they use each month.
\\
\hline
\textbf{60k fine-tune}
&
Google is already the world's most popular phone software provider, and a pay-TV operator - and now it wants to be your mobile network carrier.
\newline\newline
The company has unveiled a U.S. wireless service that switches between Wi-Fi and cellular networks to curb data use and keep phone bills low.
\\
\hline
\textbf{supervised}
&
Google launched a new U.S. wireless service today. Called \textquotesingle{}Project Fi,\textquotesingle{} it switches between Wi-Fi and cellular networks. The service, called \textquotesingle{}Project Fi,\textquotesingle{} debuted today. About two months after Google announced its plans to expand its ever-growing empire into providing wireless connections for smartphones.
\\
\hline
\textbf{supervised + 60k fine-tune}
&
Google launched a new U.S. wireless service today that switches between Wi-Fi and cellular networks. The service, called \textquotesingle{}Project Fi,\textquotesingle{} debuted today, about two months after Google revealed its plans to expand its ever-growing empire into providing wireless connections for smartphones.
\\
\hline


\end{tabular}
\caption{Random ($T=0.5$) summaries for our CNN/DM summarization task, with contexts (shown truncated) randomly sampled from CNN/DM test set.}
\label{tab:samples-summarization-cnndm}
\end{center}
\end{table*}

\begin{table*}[t]
\begin{center}
\caption*{Samples for CNN/DM summarization task (part 2)}
\begin{tabular}{|p{2cm}|p{12cm}|}

\hline
    
\hline
\textbf{context article}
&
\multicolumn{1}{|C{12cm}|}{ 
With the date of their hugely anticipated bout less than a month away, Floyd Mayweather and Manny Pacquiao's training intensity has hit gruelling levels ahead of the clash in Las Vegas.
\newline\newline
Mayweather is as renowned for showing off his extravagant lifestyle as he is for going undefeated in his professional career.
\newline\newline
But the latest pictures released from his training camp, appearing to show him hanging a weight from his head, displays the extreme level he hits in order to be as fit as possible for his fights.
\newline\newline
Manny Pacquiao (right) poses alongside actor Mark Wahlberg, who visited the Filipino at training
\newline\newline
Floyd Mayweather released two pictures from his training camp appearing hold a weight from his head
\newline\newline
Mayweather (left) takes on Pacquiao in a £160million mega-fight in Las Vegas on May 2
\newline\newline
Mayweather (left) appeared to take his training intensity to a new height before his meeting with Pacquiao
\newline\newline
As for Pacquiao, he had a famous visitor check up on his preparations for May 2.\newline\newline[...\textit{Truncated}...]\newline\newline Pacquiao and Mayweather's clash will be a reported £160million mega-fight, with Pacquiao seeking to inflict a first ever defeat upon Mayweather.
\newline\newline
The 36-year-old Pacquiao also uploaded a picture to his Twitter page away from training with his daughter.
}
\\
\hline
\textbf{reference summary}
&
Manny Pacquiao took time out from training to meet Mark Wahlberg. Floyd Mayweather shared a picture of him holding a weight with his head. The due meet in Las Vegas in a £160million encounter on May 2.
\\
\hline
\hline

\textbf{zero-shot}
&
Mayweather and Pacquiao are going to clash in Las Vegas on May 2.
\\
\hline
\textbf{60k fine-tune}
&
With the date of their hugely anticipated bout less than a month away, Floyd Mayweather and Manny Pacquiao's training intensity has hit gruelling levels ahead of the clash in Las Vegas.
\newline\newline
Mayweather is as renowned for showing off his extravagant lifestyle as he is for going undefeated in his professional career.
\\
\hline
\textbf{supervised}
&
Floyd Mayweather and Manny Pacquiao will fight in Las Vegas on May 2. Pacquiao has a visitor check up on his preparations for the bout. Mayweather released pictures of his training camp to social media.
\\
\hline
\textbf{supervised + 60k fine-tune}
&
Floyd Mayweather and Manny Pacquiao will fight in Las Vegas on May 2. Mayweather released two pictures from his training camp appearing hold a weight from his head. The 36-year-old boxer is as renowned for showing off his extravagant lifestyle as he is for going undefeated in his professional career.
\\
\hline


\end{tabular}
\caption{Random ($T=0.5$) summaries for our CNN/DM summarization task, with contexts (shown truncated) randomly sampled from CNN/DM test set.}
\label{tab:samples-summarization-cnndm2}
\end{center}
\end{table*}

\begin{table*}[t]
\begin{center}
\caption*{Samples for CNN/DM summarization task (part 3)}
\begin{tabular}{|p{2cm}|p{12cm}|}

\hline
    
\hline
\textbf{context article}
&
\multicolumn{1}{|C{12cm}|}{ 
The Solar Impulse 2, the experimental plane attempting to fly around the world without using a drop of fuel, has been grounded by the weather in China.
\newline\newline
What was supposed to be an overnight pit stop in the southwestern city of Chongqing has now stretched into a two-and-a-half week stay.
\newline\newline
It's an unexpected layover that is testing the patience of the pilots and more than 60 team members, just a quarter of the way along a journey that will cover some 35,000 kilometers (21,748 miles) over five months.
\newline\newline
"You can believe that your will can influence the weather," says pilot Bertrand Piccard, who comes from a long family line of explorers.  But "then you'll be really disappointed because it doesn't work.
\newline\newline
"Or you just know that it's an adventure you do with the forces of nature."
\newline\newline
Piccard, along with Andre Borschberg, a Swiss engineer, are taking turns flying the single-seater aircraft, which took off from Abu Dhabi on March 9 and has successfully flown through Oman, India, and Myanmar.
\newline\newline
There is much at stake.\newline\newline[...\textit{Truncated}...]\newline\newline "Our boss is the sun," says Solar Impulse spokeswoman Claudia Durgnat.
\newline\newline
Durgnat says there may be a tiny window ahead, with the next possible departure from Chongqing on Tuesday. "It's not good before, and the days after don't look very clear."
}
\\
\hline
\textbf{reference summary}
&
Solar plane attempting to be first to circumnavigate world without using fuel is stuck in China. Solar Impulse 2 attempts to prove the power of renewable energy.\\
\hline
\hline

\textbf{zero-shot}
&
The plane has been grounded in China for two weeks because of bad weather.
\\
\hline
\textbf{60k fine-tune}
&
The Solar Impulse 2, the experimental plane attempting to fly around the world without using a drop of fuel, has been grounded by the weather in China.
\newline\newline
What was supposed to be an overnight pit stop in the southwestern city of Chongqing has now stretched into a two-and-a-half week stay.
\\
\hline
\textbf{supervised}
&
Solar Impulse 2 has been grounded by the weather in China. The plane took off from Abu Dhabi on March 9. The plane is trying to prove the power of renewable energy.
\\
\hline
\textbf{supervised + 60k fine-tune}
&
Solar Impulse 2 has been grounded in China for two-and-a-half weeks. The plane is attempting to fly around the world without using a drop of fuel. The team, which includes Bertrand Piccard, is taking turns flying the single-seater. The plane took off from Abu Dhabi on March 9 and has successfully flown through Oman.
\\
\hline


\end{tabular}
\caption{Random ($T=0.5$) summaries for our CNN/DM summarization task, with contexts (shown truncated) randomly sampled from CNN/DM test set.}
\label{tab:samples-summarization-cnndm3}
\end{center}
\end{table*}
    

\end{document}